\documentclass{article}
\usepackage[preprint]{icml2026}
\usepackage{graphicx}        
\usepackage{amsmath}        
\usepackage{amssymb}        
\usepackage{hyperref}       
\usepackage{xcolor}         
\usepackage{listings}       
\usepackage{courier}        
\usepackage{placeins}       
\usepackage[toc,page]{appendix} 
\usepackage{subcaption}   % <- defines the {subfigure} environment
\usepackage{booktabs}
\usepackage{multirow}
\usepackage{float}

\icmltitlerunning{Late-Stage Generalization Collapse in Grokking}

\begin{document}
\raggedbottom
\twocolumn[
\icmltitle{Late-Stage Generalization Collapse in Grokking:\\
Detecting anti-grokking with WeightWatcher}

\begin{icmlauthorlist}
\icmlauthor{Hari K. Prakash}{ucsd}
\icmlauthor{Charles H. Martin}{calc}
\end{icmlauthorlist}

\icmlaffiliation{ucsd}{University of California San Diego, Data Science and Engineering}
\icmlaffiliation{calc}{Calculation Consulting, San Francisco, CA}

\icmlcorrespondingauthor{Hari K. Prakash}{hprakash@ucsd.edu}
\icmlcorrespondingauthor{Charles H. Martin}{charles@calculationconsulting.com}

\vskip 0.3in
]

\printAffiliationsAndNotice{} % REQUIRED
\begin{abstract}
\emph{Memorization} in neural networks lacks a precise operational definition and is often inferred from the grokking regime, where training accuracy saturates while test accuracy remains very low. We identify a previously unreported third phase of grokking in this training regime: \emph{anti-grokking}, a late-stage collapse of generalization.

We revisit two canonical grokking setups: a 3-layer MLP trained on a subset of MNIST and a transformer trained on modular addition, but extended training far beyond standard. In both cases, after models transition from pre-grokking to successful generalization, test accuracy collapses back to chance while training accuracy remains perfect, indicating a distinct post-generalization failure mode.

To diagnose anti-grokking, we use the open-source \texttt{WeightWatcher} tool based on HTSR/SETOL theory. The primary signal is the emergence of \emph{Correlation Traps}: anomalously large eigenvalues beyond the Marchenko--Pastur bulk in the empirical spectral density of shuffled weight matrices, which are predicted to impair generalization. As a secondary signal, anti-grokking corresponds to the average HTSR layer quality metric $\alpha$ deviating from $2.0$.  Neither metric requires access to the test or training data.

We compare these signals to alternative grokking diagnostic, including $\ell_2$ norms, Activation Sparsity, Absolute Weight Entropy, and Local Circuit Complexity.  These track pre-grokking and grokking but fail to identify anti-grokking. Finally, we show that Correlation Traps can induce catastrophic forgetting and/or prototype memorization, and observe similar pathologies in large-scale LLMs, like OSS GPT 20/120B.
\end{abstract}

\section{Introduction}
\begin{figure}[ht]
    \centering
    \includegraphics[width=0.40\textwidth]{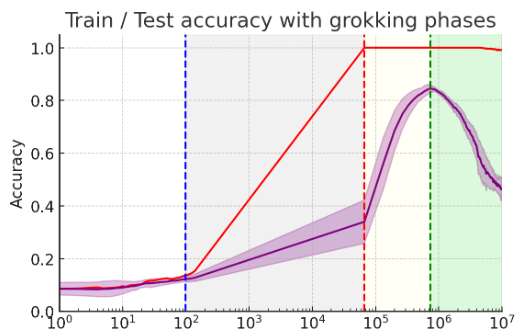}
    \caption{\textbf{The three phases of grokking.}  Training curves for a depth-3, width-200 MLP on MNIST. The initial \textbf{pre-grokking} phase (grey): training accuracy (red line) surges at $10^2$ steps, saturating between $10^4-10^5$ steps, while test accuracy (purple line) remains low; the \textbf{grokking} phase (yellow):  with test accuracy rapidly increasing after $\sim 10^5$ steps, and reaching a maximum at $10^6$ steps; and the newly revealed late-stage \textbf{anti-grokking} phase (green): test accuracy collapses (to $0.5$).}
    \label{fig:training_curves}
\end{figure}

Grokking is an intriguing phenomenon where a neural network achieves near-perfect training accuracy quickly, yet the test accuracy lags significantly, often near chance level, before abruptly surging towards high generalization \cite{power2022grokking}. Figure~\ref{fig:training_curves} illustrates this for a depth-3, width-200 ReLU MLP trained on a subset of MNIST. 
\begingroup\makeatletter
\@ifpackagewith{icml2026}{preprint}{%
  \footnote{This paper is a substantially extended version of our earlier ICML workshop paper~\cite{prakash2025grokking}.}%
}{}%
\endgroup

To dissect this phenomenon and uncover deeper dynamics, we use the open-source \texttt{WeightWatcher} tool (\texttt{WW}), which implements layer quality metrics derived from theory 
(HTSR, SETOL).\cite{martin2025setol,martin2021implicit,martin2021predicting}  The tool examines the empirical spectral density (ESD) of individual layer weight matrices $(\mathbf{W})$, providing insight into both the correlation structure, which is associated with good generalization, and statistical atypicalities, which hurt the model.  Specifically, we monitor $2$ \texttt{WW} metrics, the number of \emph{Correlation Traps}, and the heavy-tailed power law (PL) exponent $\alpha$. 

To study grokking, we reproduce two classic experiments
\cite{liu2023omnigrokgrokkingalgorithmicdata,nanda2023progress}:
 (i) a 3-layer MLP trained on a subset of MNIST (with weight decay, \texttt{WD=0}, and without, \texttt{WD>0}) and
 (ii) a small transformer trained to learn modular addition (MA)
 We run both, however,  significantly longer than typically examined. We observe that after each model has transitioned from pre-grokking to grokking (successful generalization), test accuracy collapses back to chance while training accuracy remains perfect. We denote this new phase \textbf{anti-grokking}. Superficially, anti-grokking resembles pre-grokking (perfect train / poor test).  But they are very different, both from each other and in different models.

Using the \texttt{WW} tool, we can readily identify all three phases of grokking directly from the layer ESDs.  

First, the anti-grokking phase is strongly associated with the presence of numerous Correlation Traps in both models, exactly as predicted by theory\cite{martin2025setol}.

Second, the layer $\alpha$  distinguishes the kind of memorization seen in the pregrokking phase in each model, tracks the transition into the grokking phase in both, and is correlated with a subsequent decrease in generalization in the anti-grokking phase.   In the MLP model, $\alpha$ is a  secondary but clear signal; in the MA case, however, because the model overfits the data in all phases, the  $\alpha$ signal is distorted and one needs to also inspect the layer ESDs visually.

For comparative context, we also investigate several other methodologies:
% move this to the MLP studies
\begin{enumerate}
    \item \textbf{Weight Norm Analysis:} Motivated by studies like Liu et al. \cite{liu2023omnigrokgrokkingalgorithmicdata}, we examine the $\ell_2$ norm of the weights. We observe that grokking occurs even without weight decay (leading to an increasing norm), suggesting weight norm alone is not a complete explanation, confirming subsequent findings.  \cite{golechha2024progressmeasuresgrokkingrealworld} 
    \item \textbf{Grokking Progress Measures:} We utilize metrics proposed by Golechha \cite{golechha2024progressmeasuresgrokkingrealworld}. Activation Sparsity, Absolute Weight Entropy, and Approximate Local Circuit Complexity, which capture broader structural and functional changes in the network during training.
\end{enumerate}

% Our central finding is twofold: First, HTSR, specifically the $\alpha$ exponent, not only correlates with the grokking transition but uniquely predicts a subsequent dip in test performance. Second, by extending training significantly longer than typical studies ($>10^6$ steps), we reveal a novel "ungrokking" phase—a dramatic collapse in generalization—which is also foreshadowed by the $\alpha$ metric, unlike the other measures considered. This highlights the power of HTSR in probing the fine-grained, long-term learning dynamics of neural networks.

% add one part about the types of memorization.

\textbf{Our Contributions:}
Our work makes several related contributions that help explain the underlying mechanisms associated with the grokking phenomenon:

\begin{enumerate}   
%\item By extending training significantly (up to $10^7$ steps) under zero weight decay \texttt{WD=0}, we identify and characterize \textbf{late-stage generalization collapse}: a substantial drop in test accuracy long after initial grokking, despite perfect training accuracy and a continually increasing $\ell_2$ weight norm.  We call this \textbf{anti-grokking.} % weight decay is dependant on which one we did.

\item We identify different types of 'memorization' in the pre-grokking phases.  In the MLP case, we find each layer shows a weakly-correlated type of memorization, similar to observations of pretrained models, where $\alpha > 2$.  In stark contrast, with MA (modular addition), the very-heavy-tailed (VHT) PL structure extends beyond the tail to the whole ESD, complicating $\alpha$ fits, and indicating that the training data is very-strongly-memorized.

\item At peak grokking, when test accuracy is maximized, in both  models, the \emph{average} $\alpha\approx 2$, as predicted by theory, but with some noise in the estimate.  

\item By extending training significantly for both models, we identify and characterize \textbf{late-stage generalization collapse}: a substantial drop in test accuracy long after grokking, and the presence of numerous Correlation Traps.  We call this \textbf{anti-grokking.}

\item In the MLP model, the addition of non-zero weight decay, \texttt{WD>0}, suppresses the onset of Correlation Traps and 
the extent of the test accuracy drop.

 %  \item The HTSR layer quality metric $\alpha$   effectively tracks the grokking transition under both the traditional setting of weight decay \texttt{WD>0} and zero weight decay \texttt{WD=0}, outperforming the $\ell_2$ weight norm and the other progress metrics. The HTSR $\alpha$ can distinguish between all 3 phases of grokking.
   
    \item In the MLP model, we identify the mechanism of the pre-grokking phase. Only a subset of the model layers are well trained (i.e. $\alpha \le 4$),  whereas at least one layer is underfit (i.e. $\alpha \ge 5$). Moreover, the layers can show great variability between training runs, indicating their instability.  Importantly, the layer $\alpha$'s here are distinct from those in the anti-grokking phases.  We call this \textbf{weakly-correlated memorization}.
      
    \item  Anti-grokking is identified with the average layer $\alpha$ significantly deviating from $\alpha\approx 2$. In the MLP model, anti-grokking is identified with one or more the layer $\alpha<2$, signifying a new kind of overfitting, and as predicted by theory.  
    In the MA model, however, since all the layers are overfit already, the anti-grokking is associated with an increase in $\alpha>2$, signifying \emph{catastrophic forgetting}.

    \item In the MLP model, we show how specific Correlation Traps correspond to memorizing a specific prototype instance of the training data, leading to the model being 'confused'; we call this \textbf{prototype overfitting}.  

    \item In the MA model, at peak grokking, the ESDs are fully PL across the entire density (indicating strong overfitting) and have just one or a few large eigenvalues,  as opposed to a fully formed PL tail.   This is unique to the MA model; we call this \textbf{rule-based memorization}.

%Also, in this phase, we observe the presence of anomalous rank-one (or greater) perturbations in one or more underlying layer weight matrices $\mathbf{W}$. We call these \textbf{Correlation Traps} and identify them by randomizing $\mathbf{W}$ elementwise, forming $\mathbf{W}^{rand}$, and looking for unusually large eigenvalues, $\lambda_{trap} \gg \lambda^{+}$ (where $\lambda^{+}$ is the right-most edge of the associated Marchenko-Pastur (MP) distribution \cite{martin2025setol}).

    % average alpha, highlight correlation trap and secondary observation average alpha. Correlation traps predicted by theory (SETOL) rather than redefining them.
    
\end{enumerate}

\section{Related Work}
\label{sec:related_work}

Grokking \cite{power2022grokking} is the surprising delayed emergence of generalization well after training accuracy saturates; there is  significant research into its underlying mechanisms. Initial studies often explored grokking in algorithmic tasks, frequently linking the phenomenon to the presence of weight decay (WD) which favors simpler, lower-norm solutions \cite{liu2023omnigrokgrokkingalgorithmicdata}. Other approaches include mechanistic interpretability \cite{nanda2023progress} and analyses identifying competing memorization and generalization circuits \cite{varma2023explaining, merrill2023tale}.

Varma et al. defined 'ungrokking' as generalization loss when retraining a grokked network on a \emph{smaller} dataset ($D < D_{crit}$), attributing it to shifting circuit efficiencies under WD. In contrast, we observe \textbf{late-stage generalization collapse} ('anti-grokking') occurring on the \emph{original} dataset after prolonged training $\sim10^7$ \emph{without} WD (\texttt{WD=0}). This distinct phenomenon is not predicted as it falls outside of the crucial weight decay assumption on which it relies.

Grokking studies now include real-world tasks \cite{golechha2024progressmeasuresgrokkingrealworld, humayun2024deep}. Golechha introduced progress measures (e.g., Activation Sparsity, Absolute Weight Entropy) and notably observed grokking without \texttt{WD}, resulting in increasing $\ell_2$ norms, similar to our setup. We use their metrics for comparison but extend training drastically (up to $10^7$ steps), revealing the subsequent 'anti-grokking' collapse—a phenomenon not reported in their work (even with \texttt{WD=0}).

% From the theory of Heavy-Tailed Self-Regularization (HTSR) \cite{martin2021implicit, martin2021predicting}, we track the PL exponent $\alpha$. We find $\alpha$ predicts both the initial grokking and, uniquely, the subsequent dip and eventual anti-grokking collapse with \texttt{WD=0}. Here, we identify and characterize this anti-grokking phenomenon, extending prior work that either required WD or did not explore sufficiently long training horizons.

\vspace{-0.25cm}
\section{\texttt{WeightWatcher} Layer Quality Metrics}
\label{sec:measures_metrics}
The open-source  \texttt{WeightWatcher} tool (\texttt{WW}) implements several metrics described by both the early HTSR theory and the more recent SETOL theory. \cite{martin2021implicit,martin2021predicting,martin2025setol}  They are based on examining the individual layer spectral densities using techniques adapted from Random Matrix Theory (RMT),
and rigorously established by Statistical Mechanics and extensive experimental observations.

\paragraph{Empirical Spectral Density (ESD)}
For each layer weight matrix $\mathbf{W}\in\mathbb{R}^{N\times M}$, form the \emph{correlation} matrix
%\[
\begin{equation}
\mathbf{X}\;=\;\frac{1}{N}\,\mathbf{W}^{\top}\mathbf{W}\;\in\;\mathbb{R}^{M\times M}.    
\end{equation}

Let $\{\lambda_i\}_{i=1}^{M}$ be the eigenvalues of $\mathbf{X}$.
The ESD is the discrete measure
\begin{equation}
    \rho_{emp}(\lambda)\;=\;\frac{1}{M}\sum_{i=1}^{M}\delta\!\bigl(\lambda-\lambda_i\bigr).
\end{equation}
% \rho_{emp}(\lambda)\;=\;\frac{1}{M}\sum_{i=1}^{M}\delta\!\bigl(\lambda-\lambda_i\bigr).

\paragraph{Random Matrix Theory (RMT)}
If the entries of $\mathbf{W}$ are i.i.d.\ (and well behaved), then, in the limit $N\!\to\!\infty$,$M\!\to\!\infty$  with aspect ratio $Q=N/M\ge1$ fixed, $\rho_{emp}(\lambda)$ converges to the Marchenko–Pastur (MP) density \cite{marchenko1967distribution}
%\[
\begin{equation}
\rho_{\mathrm{MP}}(\lambda)=
\begin{cases}
\displaystyle
\frac{Q}{2\pi\sigma^{2}}
\frac{\sqrt{(\lambda^{+}-\lambda)(\lambda-\lambda^{-})}}{\lambda},
& \lambda\in[\lambda^{-},\lambda^{+}],\\
0, & \text{otherwise},
\end{cases}
\end{equation}
\begin{equation}
\lambda^{\pm}=\sigma^{2}\bigl(1\pm Q^{-1/2}\bigr)^{2}.
\end{equation}

The MP distribution provides a ``null model’’ against which real, trained weight matrices can be compared.

Generally speaking, the layers in a well-trained NN, and even those in a poorly trained one, rarely follow the MP RMT theory.  
The \texttt{WW} tool looks at each layer weight matrix and characterizes the deviations from the MP RMT distribution, providing both layer quality metrics and visualizations of the ESDs that can be used to analyze the generalization performance of trained NNs.
Here, we examine two layer quality metrics: (i) the HTSR PL exponent $\alpha$, and (ii) the SETOL Correlation Traps.

\subsection{Heavy–Tailed Self-Regularization (HTSR)}
\label{subsec:HTSR}

Prior work shows that the ESD of real-world DNN layers with learned correlations almost never sits entirely within the MP bulk; instead, the right edge decays into a power law (PL) tail. Formally,

\begin{equation} \label{eq:PL_model}
\rho_{emp}(\lambda)\;\sim\;\lambda^{-\alpha},
\qquad
\lambda_{\min}<\lambda<\lambda_{\max},
\end{equation}

with the exponent $\alpha$ quantifying the strength of the correlations. According to the HTSR framework, different ranges of $\alpha$ corresponds to different phases of training and different levels of convergence for each layer:
\begin{itemize}
    \item \(\alpha \gtrsim 5-6\): \textbf{Random-like or Bulk-plus-Spikes}  the spectrum is close to the MP baseline; little task structure is present. 
    \item \(2 \lesssim \alpha \lesssim 5-6\): \textbf{Weak (WHT) to Moderate Heavy (Fat) Tailed (MHT)}  Correlations build up; layers are well-conditioned and typically generalize better.
    \item \( \alpha = 2\) \textbf{Ideal value:} Corresponds to fully optimized layers in models.  Associated with layers in models that generalize best.
    \item \(\alpha < 2\): \textbf{Very-Heavy-Tailed (VHT)} Extremely heavy tails indicate overfitting to the training data and often precede and/or are associated with decreases in the generalization / test accuracy.  For $\alpha\ll 2$, the PL fit may extend past the tail to the entire ESD.
\end{itemize}
The lower bound $\alpha=2$ on the Fat-Tailed phase is a hard cutoff, whereas the upper bound $\alpha\gtrsim 5-6$ is looser because it can depend on the aspect ratio $Q$. The presence of anomalies (i.e Correlation Traps, etc.), however, can interfere with the PL fits of $\alpha$.

\paragraph{Theoretical justification of $\alpha$:} This is outlined in a new theory, complementary to HTSR, denoted SETOL: \emph{Semi-Empirical Theory of Learning}.\cite{martin2025setol}
SETOL posits that the individual layers of a NN will come to thermodynamic equilibrium when the model obtains 
its optimal generalization accuracy
for its training data set.
The HTSR PL exponent $\alpha$ then describes the individual contribution a NN layer makes to the total model generalization (called the \emph{Layer Quality}, and where smaller  $\alpha$ is better).
Importantly, the condition $\alpha=2$
aligns with an optimal state
analogous to a phase boundary between
ideal generalization and overfitting (as explained by Wilson's  Exact Renormalization Group).
A layer with $\alpha<2$, and/or one or more Correlation Traps (below), corresponds to states of overfitting, as observed here. 

\paragraph{Estimating \texorpdfstring{$\alpha$}{alpha}.}
Following \cite{martin2021predicting}, we fit the tail of $\rho_{emp}$ to a PL \ref{eq:PL_model} through the maximum likelihood estimator (MLE) \cite{clauset2009power}.
The start of the PL tail, $\lambda_{\min}$, is chosen to minimize the Kolmogorov-Smirnov distance $D_{KS}$ between the empirical and fitted distributions.
All calculations are performed with \texttt{WeightWatcher} v0.7.5.5 \cite{weightwatcher}, which automates:
\vspace{-0.4\baselineskip}
\begin{itemize}
  \setlength{\topsep}{0pt}
  \setlength{\partopsep}{0pt}
  \setlength{\itemsep}{0.05\baselineskip}
\item SVD for the singular values $\sigma_i$ ($\lambda_i=\sigma_i^2$),
\item PL fits and $D_{KS}$ tests (including $\lambda_{\min}$ and $\lambda_{\max}$),
\item Detecting Correlation Traps.
\end{itemize}
\vspace{-0.3\baselineskip}

%\begin{figure}[h!]
%    \centering
%    \includegraphics[width=0.6\textwidth]{PL-Fit.png} 
%    \caption{Example of a power-law fit to the tail of the Empirical Spectral Density (ESD) for a network %layer (e.g., Layer 2 at a specific training step). The red line indicates $\lambda_{min}$, and the dashed %red line shows the fitted power law with exponent $\alpha$. The plot title includes the estimated %$\alpha$, the KS distance $D_{KS}$, $\lambda_{min}$, and the ESD sigma $\sigma$.}
%    \label{fig:pl_fit_example_main} % Renamed from fig:pl_fit_example for clarity if needed, or use %original
%\end{figure}

\begin{figure}[t]
  \centering
  \begin{minipage}[b]{0.48\columnwidth}
    \centering
    \includegraphics[width=\linewidth]{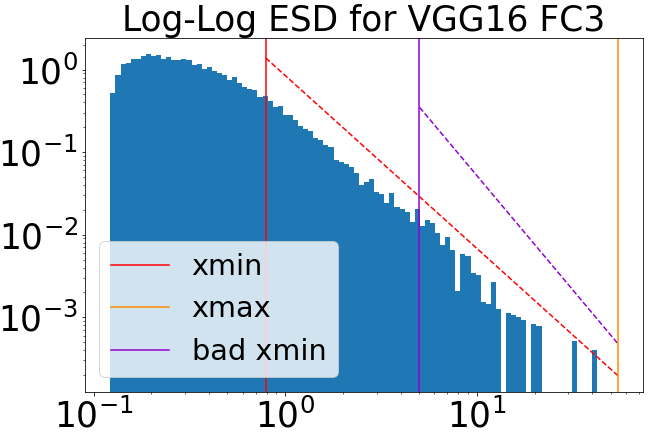}
  \end{minipage}\hfill
  \begin{minipage}[b]{0.48\columnwidth}
    \centering
    \includegraphics[width=\linewidth]{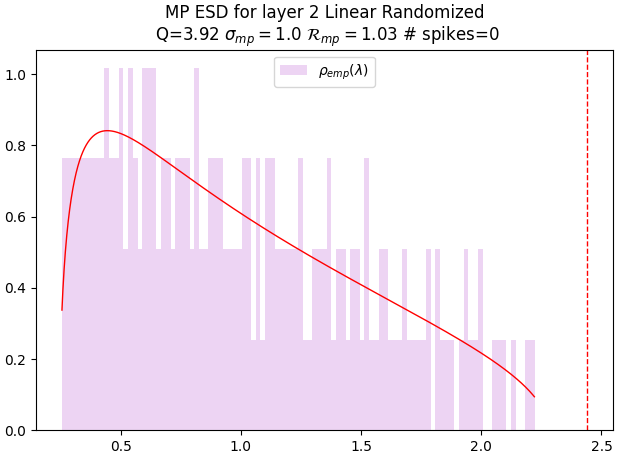}
  \end{minipage}
  \caption{\textbf{Left}: Example of the ESD derived from a well-correlated $\mathbf{W}$ (blue) with a power-law tail fit (red), on a \textbf{log–log} plot.
  \textbf{Right}: Example of the ESD of $\mathbf{W}^{\mathrm{rand}}$ (light purple) with a Marchenko–Pastur (MP) fit (red), shown on a \textbf{log–linear} plot.}
  \label{fig:ESDs}
\end{figure}

Figure~\ref{fig:ESDs} (Left) shows a PL fit on a log-log scale for a representative layer after training. The plot displays the ESD for a typical NN layer  (a histogram or kernel density estimate of $\rho_{emp}(\lambda)$), the automatically chosen $\lambda_{\min}$ (xmin, vertical line, red),  the $\lambda_{\max}$ (xmax, vertical line, orange), and the best fit for the PL tail (dashed line, red).  
%The estimated $\alpha$ value and the $D_{KS}$ statistic are usually reported on such plots.  
%\FloatBarrier 

The PL fit is very sensitive to the choice of $\lambda_{\min}$, and a poor choice will result in a poorly estimated $\alpha$. If $\lambda_{\min}$ is too large (bad xmin, vertical line, purple), then the PL tail is too small and results in a larger $\alpha$. Selecting $\lambda_{\min}$ defines the tail alpha $(\alpha)$ and is fully automated using the open-source \texttt{WeightWatcher} tool.\cite{weightwatcher}

Note that in MA model (see Fig.~\ref{fig:modadd-embed-esd-panel}), it is very hard to get a reliable PL fit to the tail because the entire ESD is clearly VHT PL, with an effective (full ESD) PL exponent $\alpha\ll 2$.

\paragraph{Significance for Grokking/Anti-Grokking.}
Across the MLP experiments, the trajectory $\alpha(t)$ proves to be a highly sensitive indicator of the network’s generalization state: large drops toward $\alpha\!\approx\!2$ coincide with the onset of grokking, while a further fall below $\alpha < 2$ foreshadows (and then characterizes) the "anti-grokking" collapse. 

% \vspace{2mm}
% \noindent
% \textbf{References}\;(\emph{selection})\;
% \cite{marchenko1967distribution,martin2021implicit,martin2021predicting,clauset2009power,weightwatcher}
\subsection{SETOL and Correlation Traps}
The SETOL theory predicts that the onset of  \textbf{Correlation Traps} will induce a phenomenon like  anti-grokking (generalization collapse).  Such traps are specific atypicalities in the  weight matrix elements.  What are they and how are they detected using \texttt{WeightWatcher}?

If we randomize a layer weight matrix elementwise, 
\begin{equation}
\mathbf{W}\rightarrow\mathbf{W}^{rand},
\end{equation}
 we then expect the ESD of $\mathbf{W}^{rand}$ to be well-fit by an MP distribution because the elements $W_{i,j}$  should now be uncorrelated and distributed randomly; See Figure~\ref{fig:ESDs} (Right).
 SETOL argues that deviations from this are significant and informative.  To identify these deviations, 
we compare the randomized layer ESDs against the MP distribution at the different stages of training to assess deviations from randomness. We identify these deviations as anomalously large eigenvalues in the underlying $\mathbf{W}^{rand}$.  Such eigenvalues are called Correlation Traps, $\lambda_{trap}$, when they extend well beyond the bulk edge $\lambda^{+}_{rand}$ of the best fit MP distribution:
\begin{equation} \label{eq:correlation_trap} % Optional: added a label for referencing
    \text{Correlation Trap: } \qquad \lambda_{trap}\gg\lambda^{+}_{rand}
\end{equation}
For more details, see the Appendix~\ref{app:stats_of_traps}, and the SETOL monograph
\cite{martin2025setol}.
% $\lambda_{trap}$, when they are significantly larger than the bulk edge $\lambda^{+}_{rand}$ of the best fit MP distribution.
% \begin{equation}
%     \lambda_{trap}\gg\lambda^{+}_{rand}
% \end{equation}
% See the Appendix~\ref{app:stats_of_traps} for additional statistical validation of the presence of such traps, as well as the Supplementary Information. Also, see \cite{martin2025setol} for more details.

\begin{figure}[t]
  \centering
  \begin{minipage}{0.48\columnwidth}
    \centering
    \includegraphics[width=\linewidth]{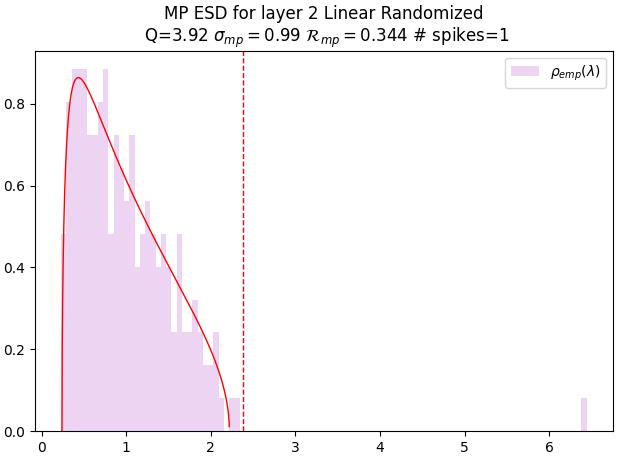}
  \end{minipage}
  \hfill
  \begin{minipage}{0.48\columnwidth}
    \centering
    \includegraphics[width=\linewidth]{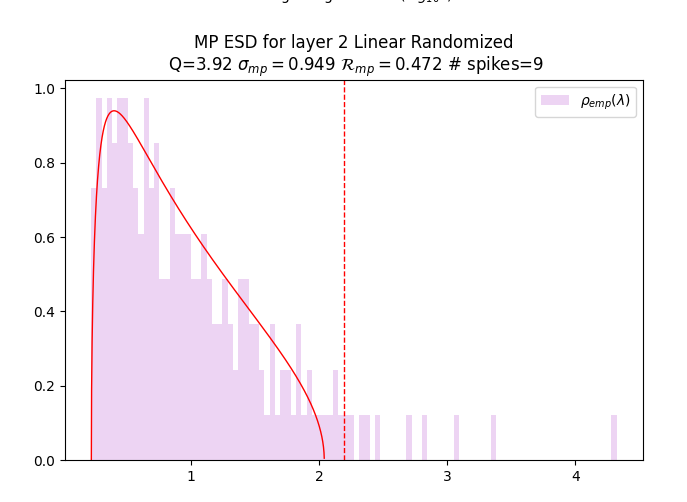}
  \end{minipage}
  \caption{\textbf{Examples of Correlation Traps}. ESDs of $\mathbf{W}^{rand}$ (light purple) of Layer~2 compared to an MP fit (red). Correlation traps $\lambda_{trap}$ appear as spikes to the right of the MP bulk (log x-axis).
  \textbf{Left: Right Before Collapse} ($\sigma_{mp}\approx0.9879$; KS $p\approx4\times10^{-13}$). A single dominant trap at $\lambda_{trap}\approx10^{6.5}$.
  \textbf{Right: Final Generalization Collapse} (KS $p\approx1.9\times10^{-5}$). Multiple traps $\lambda_{trap}\in[10^{2.x},10^{6.5}]$.}
  \label{fig:Traps}
\end{figure}

The \texttt{WeightWatcher} tool detects Correlation Traps automatically; it randomizes $\mathbf{W}$ elementwise, then performs automated MP fits by estimating the variance $\sigma^2_{MP}$ of the underlying randomized matrix $\mathbf{W}^{rand}$, finding the fit that best describes the bulk of its ESD of $\mathbf{W}^{rand}$.  It then finds all  eigenvalues $\lambda_{trap}$ that are significantly larger (i.e. beyond the Tracy-Widom fluctuations) of the MP bulk edge $\lambda^{+}_{rand}$ of the ESD of $\mathbf{W}^{rand}$. Figure~\ref{fig:Traps} depicts two layers from the models studied here with Correlation Traps.
%shrink 3 fix 14, add in correlation trap new section 4 and 5.

For additional statistical validation, here,
 we also  use the Kolmogorov-Smirnov (KS) test to quantify the dissimilarity between the  ESD of $\mathbf{W}^{rand}$ and its best MP fit.  A large difference, combined with a visual inspection of the 
 ESDs, along with a potentially lower-quality $\alpha$ fit, indicates the presence of one or more Correlation Traps.

\subsection{Other Benchmarked Metrics}
\label{subsec:other_benchmarked_metrics}

We benchmarked our MLP findings against the $\ell_2$ weight norm and several Grokking progress diagnostics proposed earlier. In total, we included:
\vspace{-0.5\baselineskip}
\begin{itemize}
  \setlength{\topsep}{0pt}
  \setlength{\partopsep}{0pt}
  \setlength{\parsep}{0pt}
  \setlength{\itemsep}{0.05\baselineskip}
\item $\ell_2$ Weight Norm ($\|\mathbf{W}\|_2$)
\item Activation Sparsity ($\Lambda_{AS}$)
\item Absolute Weight Entropy ($H_{\mathrm{abs}}(\mathbf{W})$)
\item Approximate Local Circuit Complexity ($\Lambda_{\mathrm{LC}}$)
\end{itemize}
\vspace{-0.5\baselineskip}
See Appendix~\ref{app:comparative_measures} for further discussion.

\vspace{-0.25cm}
\section{MLP Experiment: Results and Analysis} % Was Section 4, now Section 5 if Exp Setup was section 4
\label{sec:results_analysis}

Our primary signal for detecting anti-grokking is (SETOL) Correlation Traps, whereas our secondary, weaker signal is the (HTSR) $\alpha$.  We discuss the HTSR $\alpha$ results first, compare these to the other proposed grokking measures, and, finally, discuss the Correlation Traps.

\begin{figure}[!ht]
    \centering
    \includegraphics[height=0.28\textwidth]{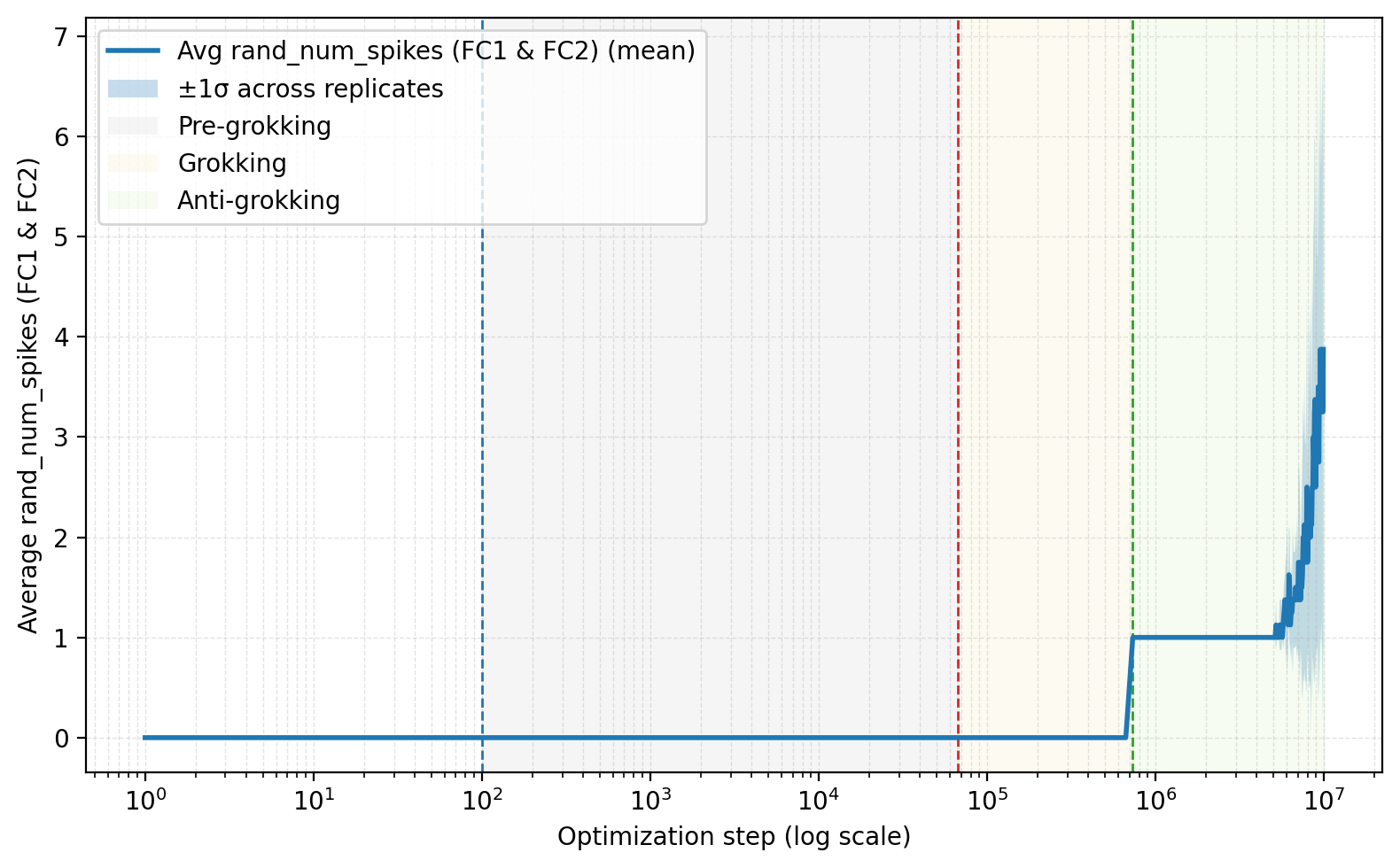}
     \caption{Average Correlation Traps across layers during optimization steps. The increase in the number of traps coincides with the "anti-grokking" performance drop seen in Fig.~\ref{fig:training_curves} at $1M$ steps. This is our primary signal for anti-grokking,}
    \label{fig:MLPtraps}
\end{figure}

\begin{figure}[!ht]
    \centering
    \includegraphics[height=0.28\textwidth]{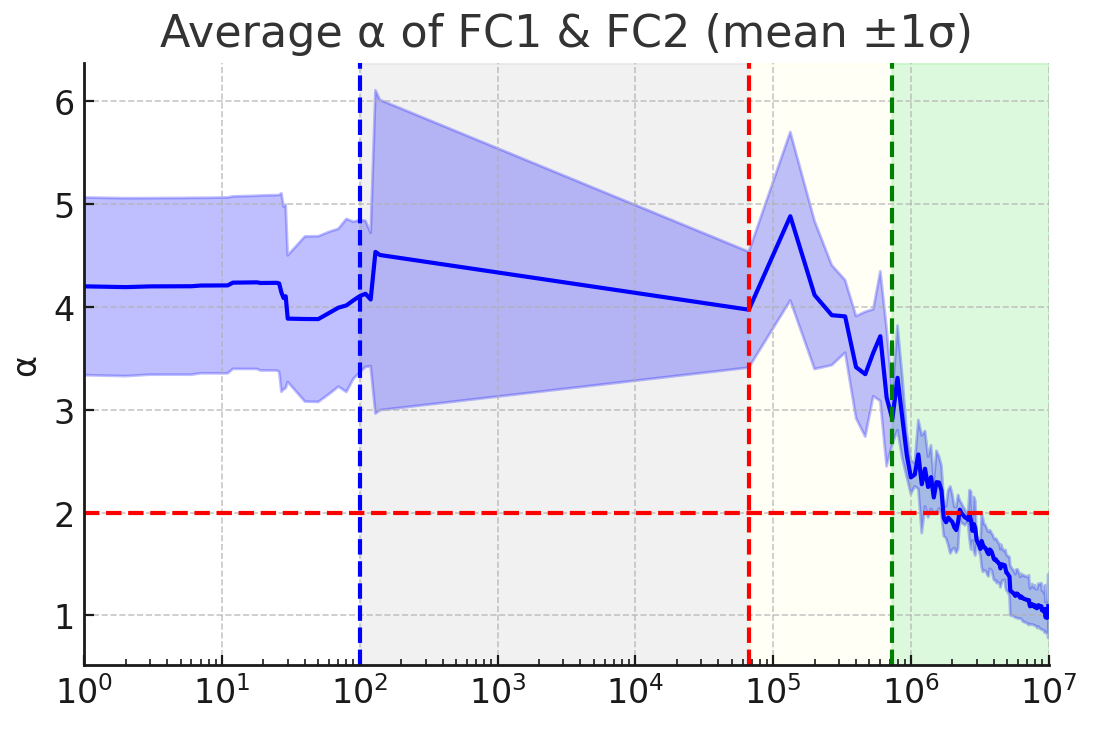}
     \caption{Average $\alpha$ across layers during optimization. Note the drop below the critical threshold $\alpha=2$, coinciding with the "anti-grokking" performance drop seen in Fig.~\ref{fig:training_curves} after $\sim 1M$ steps. This is our secondary signal for anti-grokking. From Figure~\ref{fig:replicate-alpha}.}
    \label{fig:alpha}
\end{figure}
%\FloatBarrier % Prevent figures from floating past this point

% \begin{figure}[!ht]
%     \centering
%     \includegraphics[height=0.28\textwidth]{updated_alphas_crop.png}
%      \caption{HTSR results vs. optimization steps.  Average Correlation Traps across layers. Note the increase in traps coinciding with the "anti-grokking" performance drop seen in Fig.~\ref{fig:training_curves} after ~1M steps.}
%     \label{fig:alpha}
% \end{figure}

\subsection{WeightWatcher Metrics for Tracking Grokking}

\paragraph{HTSR layer quality metric $\alpha$:}
The HTSR  $\alpha$ reveals critical dynamics missed by other measures. Figure~\ref{fig:alpha} shows the evolution of $\alpha$ averaged across layers FC1 and FC2.   

\begin{table}[!ht]
\caption{\textbf{Layer-wise and average HTSR $\alpha$ exponents. } At the right edge of each grokking phase: Pre-grokking $\sim10^5$ steps, 
 Grokking $\sim10^6$ steps, and Anti-grokking $\sim10^7$ steps,
For the zero-weight-decay (\texttt{WD=0}) experiment; values are taken from
Fig.~\ref{fig:alpha}. Various seeds are used and variability in initialization, optimizer trajectory may occur.}%

\centering
\setlength{\tabcolsep}{6pt}
\begin{tabular}{lccc}
\toprule
\textbf{Layer} & \textbf{Pre-grokking} & \textbf{Grokking} & \textbf{Anti-grokking} \\
\midrule
FC1 $\alpha$   & $5.0 \pm 0.7$ & $3.6 \pm 0.5$ & $0.9 \pm 0.4$ \\
FC2 $\alpha$   & $2.9 \pm 0.7$ & $2.3 \pm 0.2$ & $1.3 \pm 0.3$ \\
Avg.\ $\alpha$ & $4.0 \pm 0.6$ & $2.9 \pm 0.2$ & $1.1 \pm 0.3$ \\
\bottomrule
\label{tab:grokking_phases}
\end{tabular}

% \begin{tabular}{lccc}
% \toprule
% \textbf{Layer, Metric} & \textbf{Pre-grokking} & \textbf{Grokking (Max Test Acc.)} & \textbf{Anti-grokking (Collapse)} \\
% \midrule
% FC1 $\alpha$    & $4.0 \pm 1.3$ & $3.2 \pm 0.6$ & $1.0 \pm 0.40$ \\
% FC2 $\alpha$    & $4.6 \pm 0.5$ & $2.4 \pm 0.1$ & $1.4 \pm 0.24$ \\
% average $\alpha$ & $4.3 \pm 0.70$ & $2.8 \pm 0.30$ & $1.2 \pm 0.23$ \\
% \bottomrule
% \end{tabular}
\end{table}

Initially, $\alpha$ is high, reflecting random-like weights. As training progresses and the network begins to fit the training data, $\alpha$ decreases. The sharp drop towards the optimal (fat-tailed) regime ($2 \lesssim \alpha \lesssim 5-6$) coincides with grokking phase (Figure~\ref{fig:training_curves}, $10^4$-$10^5$ steps ).

As training continues, however, and into the millions of steps, $\alpha$ consistently dips below $2$, entering the VHT regime. This occurs notably in the second fully connected layer (FC2, see Figure~\ref{fig:replicate-alpha}). This drop of $\alpha<2$ indicates  a sub-optimal layer with  overly strong correlations, occurring just after  the significant drop in test accuracy, the anti-grokking phase, observed after $10^6$ steps in Figure~\ref{fig:training_curves}.

HTSR theory predicts optimal test performance when all layers reach $\alpha \approx 2$, but only late in training here. We attribute this delay to the small dataset, which requires extended optimization for full layer convergence; however, by the time $\alpha \approx 2$ is reached, Correlation Traps emerge, degrading test performance and inducing anti-grokking.

Together, these observations highlight the unique sensitivity of the HTSR $\alpha$ metric. This metric not only identifies the grokking transition but also provides an early warning for the subsequent instability and the novel anti-grokking phenomenon, highlighting  pathological correlation structures/overfitting forming deep into training. 

Here, the HTSR $\alpha$ is a weaker but important secondary signal for anti-grokking; a much stronger, more general signal is the presence of Correlation Traps, as shown in Figure~\ref{fig:MLPtraps}. 
But first, we examine other metrics from the literature.

\begin{figure}[ht]
    \centering
    \includegraphics[height=0.30\textwidth]{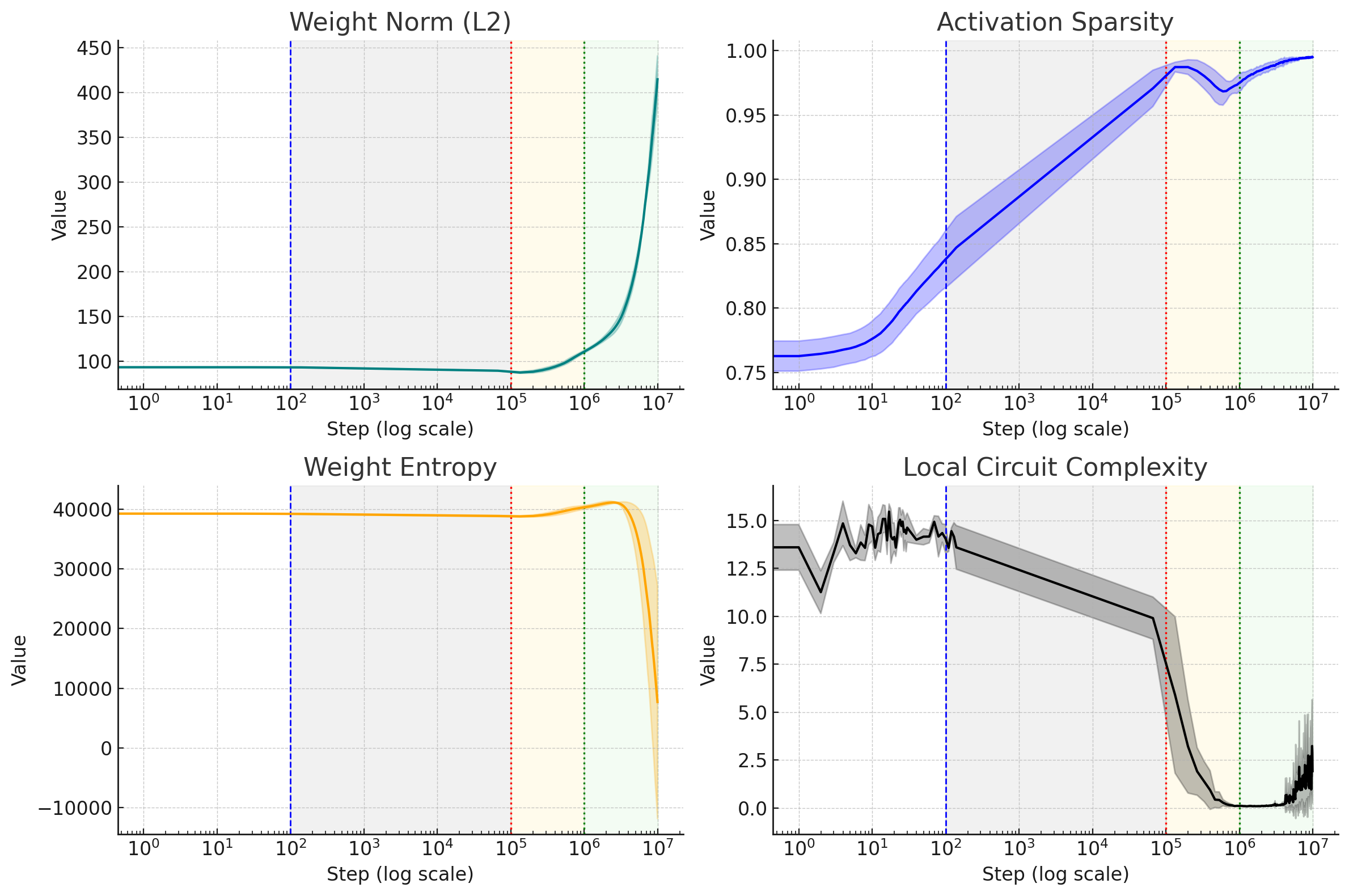}
    \caption{Alternative progress measures (Golechha \cite{golechha2024progressmeasuresgrokkingrealworld}) vs. optimization steps. Top: Activation Sparsity. Middle: Absolute Weight Entropy. Bottom: Approximate Local Circuit Complexity. While these metrics show changes during the initial training and grokking phases (Activation Sparsity for example), they do not clearly delineate onset or  magnitude of the late-stage "anti-grokking" performance dip observed after ~$10^6$ steps.}
    \label{fig:other_measures}
\end{figure}

\paragraph{Other comparative metrics:}
In contrast, the other comparative metrics capture the initial training and grokking phases but fail to predict the late-stage generalization collapse. Figure~\ref{fig:other_measures} displays the Activation Sparsity, Absolute Weight Entropy, and Approximate Local Circuit Complexity $(\Lambda_{LC})$. While these metrics show clear trends during the initial learning and grokking phases (e.g., changes in sparsity and complexity), their trajectories become relatively stable or lack distinct features corresponding to the dramatic performance drop seen during "anti-grokking". 

For example, $\Lambda_{LC}$ remains relatively flat in the late-stages up until some noise at the end, offering no warning of the impending collapse. In contrast, the onset of the collapse is immediately detected by the onset of numerous Correlation Traps, and verified (a little later) with average $\alpha<2$.

Similarly, the Activation Sparsity $(\Lambda_{AS})$ shows an inflection around just before peak test accuracy and does detect grokking, it generally continues its upward trend in the late-stage collapse. At peak test accuracy, however, it has a similar value as peak anti-grokking.  Moreover, in the control \texttt{WD>0} case, the $\Lambda_{AS}$ also shows an inflection point, but there is only very slight anti-grokking, and this just indicates that some unknown transitional change has occurred. 

%\FloatBarrier % Prevent figures from floating past this point

%%%%%
In the \texttt{WD=0} case, $\Lambda_{AS}$ generally increases throughout training (Figure~\ref{fig:other_measures}), seemingly tracking the pre-grokking and grokking phases, however, it fails the negative control in the anti-grokking phase because it continues to increase in the same way as in pre-grokking. Prior studies have linked activation sparsity to generalization \cite{li2023lazyneuronphenomenonemergence, merrill2023tale, peng2023theoretical} and reported specific dynamics such as plateauing before grokking \cite{golechha2024progressmeasuresgrokkingrealworld} or an increase preceding a rise in test loss \cite{huesmann2021impact}.  Specifically, we observe a subtle inflection or dip in $\Lambda_{AS}$ coinciding with the point of maximum test accuracy before a slight increase. While this feature appears to mark a shift around peak test accuracy, its specific predictive utility for subsequent generalization dynamics is questionable.
In other words, without knowing the proper sparsity cutoff, it is impossible to determine if increasing $\Lambda_{AS}$ corresponds to pre-grokking or anti-grokking. In  contrast, because the HTSR $\alpha=2$ is a theoretically established universal cutoff, it can distinguish the two phases.

\subsection{SETOL Correlation Traps and Anti-Grokking}

The strongest signal we have for anti-grokking is the sudden onset of Correlation Traps in both layers, in both the cases with $(WD>0)$ and without \texttt{WD=0} weight decay. As described in Section~\ref{subsec:HTSR}, we analyze the eigenvalues $\{\lambda_i\}$ of the randomized weight matrices $\mathbf{W}^{rand}$ derived from each layer's weight matrix $\mathbf{W}$ for layers FC1 and FC2.

% one blank line before the table
\begin{table}[t]
\centering
\caption{\textbf{Average \# Correlation Traps} in FC1/FC2 at the right edge of three phases:
Pre-grok ($\sim10^{5}$ steps), Grok ($10^{6}$), Anti-grok ($10^{7}$).
Without weight decay (\texttt{WD=0}) and with (\texttt{WD>0}).}
\label{tab:traps}
\small
\setlength{\tabcolsep}{3pt}
\begin{tabular}{lccc}
\toprule
\textbf{Setting, Layer} & \textbf{Pre-Grokking} & \textbf{Grokking} & \textbf{Anti-Grokking} \\
\midrule
\texttt{WD=0}, FC1   & $0\pm0$ & $0\pm0$ & $7.5\pm5.6$ \\
\texttt{WD=0}, FC2   & $0\pm0$ & $0\pm0$ & $1\pm0$ \\
\midrule
\texttt{WD>0}, FC1   & $0$     & $0$     & $2.0\pm0.0$ \\
\texttt{WD>0}, FC2   & $0$     & $0$     & $1.0\pm0.0$ \\
\bottomrule
\end{tabular}
\end{table}
First, as clearly shown in Figure~\ref{fig:MLPtraps}, for the \texttt{WD=0}  case, numerous Correlation Traps suddenly appear right at the onset of the anti-grokking phase. 
Second, as shown in Table~\ref{tab:traps}, for layers FC1 and FC2, and for both \texttt{WD} settings,  neither layer shows evidence of Correlation Traps until the anti-grokking phase.  Moreover, there are significantly more traps in the \texttt{WD=0} case, which coincides with the much stronger anti-grokking observed.
Finally, further statistical analysis for the FC2 layers is provided in Appendix~\ref{app:stats_of_traps}.  The presence of Correlation Traps, combined with $\alpha<2$ (later), is a definitive signal indicating the model is deep in the anti-grokking phase.  So what are the traps ?

\begin{figure}[t]
  \centering
  \begin{subfigure}[b]{0.32\columnwidth}
    \centering
    \includegraphics[width=\linewidth]{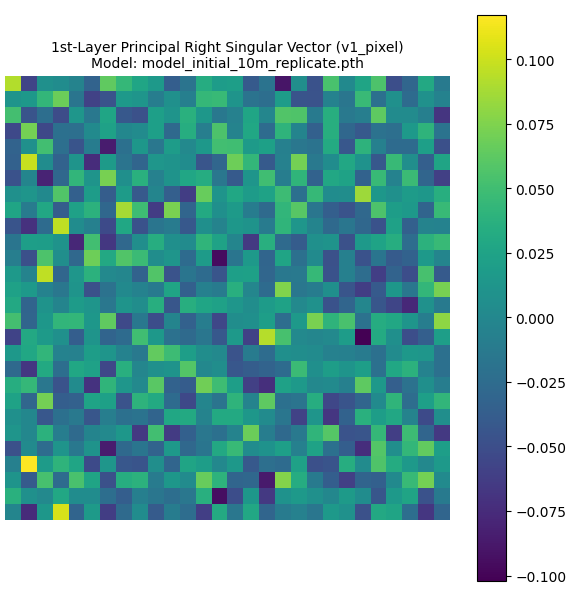}
    \caption{Pre-grokking}
  \end{subfigure}\hfill
  \begin{subfigure}[b]{0.32\columnwidth}
    \centering
    \includegraphics[width=\linewidth]{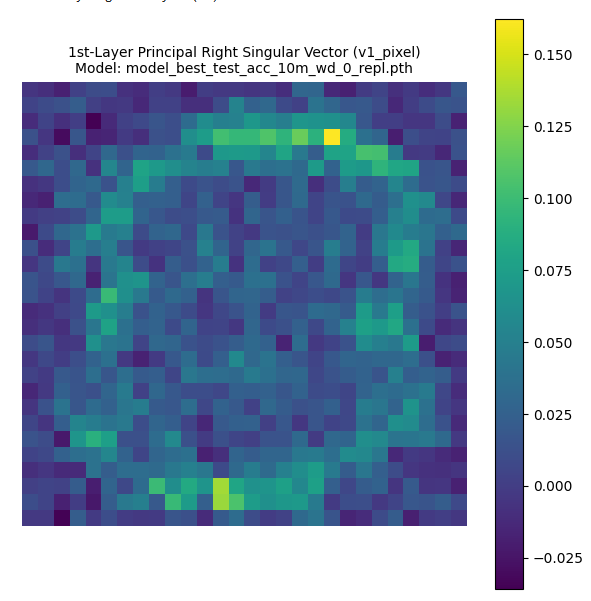}
    \caption{Grokking}
  \end{subfigure}\hfill
  \begin{subfigure}[b]{0.32\columnwidth}
    \centering
    \includegraphics[width=\linewidth]{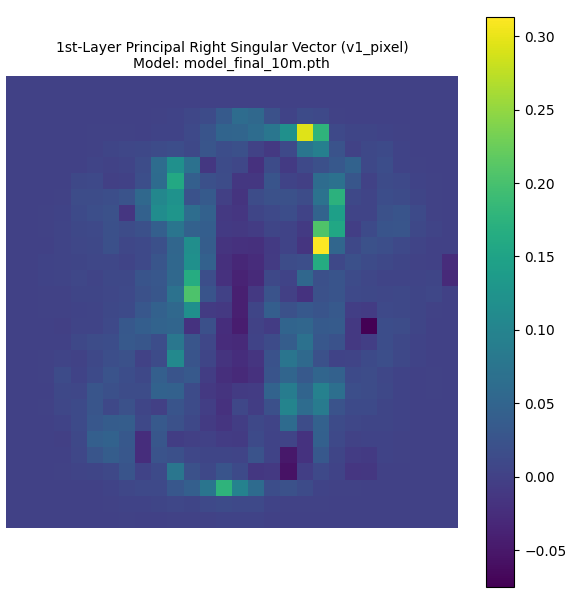}
    \caption{Anti-grokking}
  \end{subfigure}

  \caption{\textbf{Prototype-Memorization.} Largest right singular vector $v^{(1)}$ of $W_1$ in pixel space.
  $v^{(1)}$ evolves from (i) unstructured noise in pre-grokking, to
  (ii) a smooth global ring-like template during grokking, to
  (iii) localized, digit-shaped templates (resembling an ``8'') in anti-grokking.}
  \label{fig:v1-pixel}
\end{figure}

\paragraph{Correlation Traps and Prototype Memorization} We can identify certain Correlation Traps with the MLP model overfitting to specific prototypes of the training data.  As shown in Figure~\ref{fig:v1-pixel}, the leading right singular vector $v^{(1)}$ of FC1 evolves from unstructured noise (pre-grokking) to a smooth, global template at peak generalization, and finally into highly localized, digit-shaped patterns in the anti-grokking phase. These localized structures indicate that the model has begun encoding specific training prototypes at the level of individual layers, rather than learning a generalizable representation.
This is described in more detail the Appendix, Section~\ref{sec:tail_vectors}. 
Furthermore $W_1$ rows show clear digit memorization as seen in Figure~\ref{fig:l1-rfs}.
The trap(s) causes the MLP model to systematically mispredict the test examples, and is fundamentally different from the weakly-correlated overfitting in the pre-grokking phase.

\vspace{-0.25cm}
\section{MA Experiment: Results and Analysis} % Was

\begin{figure}[!ht]
    \centering
    \includegraphics[height=0.25\textwidth]{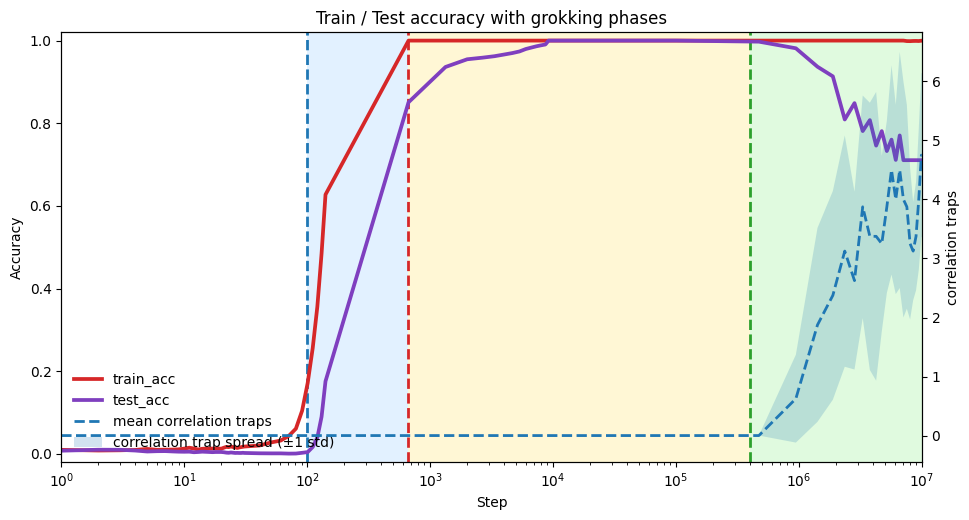}
     \caption{Average Correlation Traps across layers during optimization steps. The increase in the number of traps coincides with the anti-grokking drop in test performance (purple).}
    \label{fig:MAtraps}
\end{figure}

We summarize our results on the Modular Addition (MA) experiment below; the Appendix provides more details.

As with the MLP experiment, the MA experiment shows an anti-grokking phase when trained significantly longer than the  standard setup.  And, as with the MLP case, anti-grokking is readily detected by the onset of numerous Correlation Traps.  This is clearly seen in Figure~\ref{fig:MAtraps}, and in Table~\ref{tab:layer_traps_qn}.
Moreover, the onset of the grokking phase is also associated with the layer average HTSR PL $\alpha\approx 2$.

\begin{figure}[t]
  \centering
  \includegraphics[width=0.5\columnwidth]{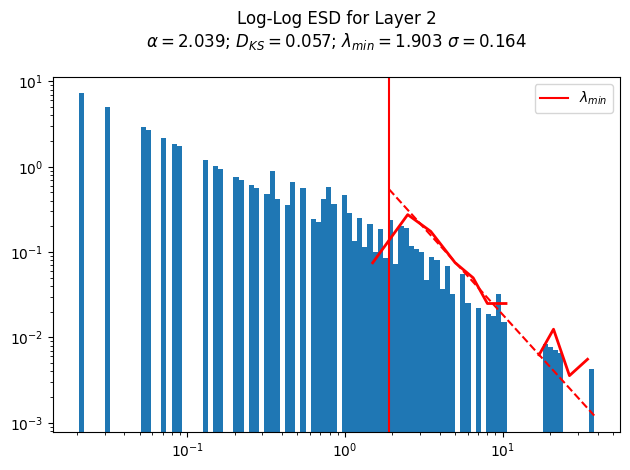}
  \caption{\textbf{Rule-based Memorization.} Grokking embedding layer ESD. Empirical spectral density (ESD) of $\mathbf{W}_{\mathrm{embed}}$ (log--log) with a PL tail fit. During grokking, the tail fit is strongest (low $D_{\mathrm{KS}}$) and $\alpha$ is near the HTSR optimum, $\alpha \approx 2$.  From Figure~\ref{fig:esd-mp-grok}.}
  \label{fig:embed-esd-grok-only}
\end{figure}

The structure of the layer ESDs, however, are significantly different from that expected from the MLP case, and indicate that in all phases nearly all layers are strongly overfit to the training data.  See Figure~\ref{fig:embed-esd-grok-only}.  Specifically, while \texttt{WeightWatcher} can fit the tail of the ESDs to a PL, and in the grokking case the PL fit  is very good ($\alpha\approx 2$, $D_{KS}=0.05$), never-the-less, the ESD in Figures~\ref{fig:modadd-embed-esd-panel} and ~\ref{fig:esd-mp-anti} are clearly VHT PL across the entire spectrum.  

Moreover, in the grokking phase, many layers have only 1 or a few very large eigenvalues, as opposed to a well formed PL tail. We call this as \textbf{rule-based memorization}.

As shown in Figure ~\ref{fig:esd-anti},  with the Correlation Traps, the PL tail shrinks and the fit reverts back to a larger (not smaller) $\alpha=3.138$.  This is typical here, and because of this, the traps appear to be inducing \emph{catastrophic forgetting.}

\vspace{-0.25cm}
\section{Conclusion} % Was Section 5, now Section 6
\label{sec:conclusion}
This study investigated the well-known grokking phenomenon in neural networks (NN) using the open-source \texttt{WeightWatcher} tool \cite{weightwatcher}.  
We study  the long-term generalization dynamics of the grokking phenomenon in an MLP model on MNIST, both with weight decay (\texttt{WD>0}) and without (\texttt{WD=0}, and in the classic modular addition (MA) experiment using a small transformer.  

We compare the application of
the tool's layer quality metrics, the HTSR $\alpha$ metric, and the presence of Correlation Traps from the more recent SETOL theory.  We track the number of traps and the layer $\alpha$ across optimization steps, and visually inspect the ESDs as needed.

For both the MLP model (notably in the \texttt{WD=0} setting), and for the MA model, we observe a novel \textbf{late-stage generalization collapse}, called \textbf{anti-grokking}. This collapse is characterized by a significant drop in test accuracy despite sustained perfect training accuracy (and a large $\ell_2$ norm)  after extensive training (up to $10^7$ steps). 

Our primary finding is that we can reliably detect this new anti-grokking phases by observing the onset of numerous Correlation Traps in one or more layers.  
The effect of such traps is predicted by the recently developed SETOL statistical mechanics theory underlying the \texttt{WW} tool.

Secondarily, we find that at or near peak grokking, as predicted by theory, the average layer $\alpha\approx 2$ in both models, as predicted, and, moreover,  the later anti-grokking phase is seen in the MLP case by 1 or more layers with $\alpha<2$.

For the MLP models, 
we compare these metrics
to the $\ell_2$ norm and  several previous proposed grokking progress measures \cite{liu2023omnigrokgrokkingalgorithmicdata,golechha2024progressmeasuresgrokkingrealworld}, and for both settings (\texttt{WD=0\;, WD>0}).
Previous work has attempted to explain grokking (using the $\ell_2$ norm), but only succeeds in the presence of weight decay (\texttt{WD}), and has been unable to explain grokking without weight decay.  
We examined 3 other grokking progress measures, plus the $\ell_2$ norm,  including Activation Sparsity $\Lambda_{AS}$, Absolute Weight Entropy $H_{abs}(W)$, and Approximate Local Circuit Complexity $\Lambda_{LC}$.  Although $\Lambda_{AS}$ and  $\Lambda_{LC}$ captured the first 2 grokking phases, and do change at the anti-grokking transition, they fail to unambiguously predict anti-grokking.

%In particular, we examine the long-term generalization dynamics in neural %networks, contrasting training under zero weight decay (\texttt{WD=0}), with training %under non-zero weight decay (\texttt{WD>0}).   Our primary finding, leveraging the %theory ofHeavy-Tailed Self-Regularization (HTSR) \cite{martin2021implicit}, is %that the heavy-tailed exponent $\alpha$ effectively tracks the initial grokking %transition and subsequent performance dips in both regimes. 
%which leads to continuously increasing $l2$ norms,
%, where $l2$ norms typically decrease or stabilize.
%Critically, in the \texttt{WD=0} setting, $\alpha$ also provides an early indication of %a novel \textbf{late-stage generalization collapse}. This collapse is %characterized by a significant drop in test accuracy despite sustained perfect %training accuracy (and a large $l2$ norm), and is observed after extensive %training (up to $10^7$ steps).

% We also examined several other grokking progress measures, including the $l2$ norm \cite{liu2023omnigrokgrokkingalgorithmicdata}, Activation Sparsity, Absolute Weight Entropy, and Approximate Local Circuit Complexity \cite{golechha2024progressmeasuresgrokkingrealworld}. 
%None of these measures captured the complete evolutionary aspects of the network in both cases (WD=0,\;WD>0), nor did they offer comparable signals for late-stage collapse (specifically for \texttt{WD=0}). 

We propose a new explanation of the grokking phenomenon. In the pre-grokking phase, where only training accuracy saturates, only a subset of layers converge ; and not fully.  Crucially, layers most relevant for generalization have not yet converged. During the grokking phase, when test accuracy is maximal, important layers converge strongly, with average $\alpha$ eventually approaching the optimal value $\alpha \approx 2.0$,  as predicted~\cite{martin2025setol}. In the anti-grokking phase, test accuracy drops because one or more layers exhibit Correlation Traps (and sometimes rank collapse) which hurt generalization.

The MLP and MA models both exhibit this grokking behavior in their layers, but the nature of memorization is very different. In the MLP model, pre-grokking memorization starts as weakly-correlated, and in the anti-grokking phase, Correlation Traps cause  overfitting to specific training data prototypes.
In the MA model, 
all of the layers are very-strongly overfit to the training data in all phases, but correlations still concentrate into a Fat PL tail at peak grokking. In anti-grokking, numerous Correlation Traps also appear, but here causing catastrophic forgetting.

We consider the implications of observing numerous Correlation Traps in the anti-grokking phase, both in the MLP and MA models.  The 'traps' are anomalous rank-one (or greater) perturbations in the weight matrix $\mathbf{W}$, causing a large mean-shift in the distribution of elements: 
$\mathbb{E}[W_{ij}]\rightarrow\textit{large}$  and distorting the ESDs and interfering with PL tail fits of $\alpha$.  The large shift in $\mathbb{E}[W_{ij}]\rightarrow\textit{large}$ indicates that the distribution of weights is \emph{atypical}. That is, different random samples of the weights could have very different means.  And as with any statistical estimator, an atypical distribution will not generalize well, irrespective of the particular mechanism of overfitting (or forgetting).

%(Similar results have been seen in training a similar model with very large learning rates\cite{martin2025setol}.) %Consequently, it is hypothesized that layers with large numbers of Correlation Traps are overfitted to the training data (in some unspecified way), and hurt the model test accuracy.

\begin{figure}[!ht]
  \centering
  \begin{minipage}[b]{0.45\columnwidth}
    \centering
    \includegraphics[width=\linewidth]{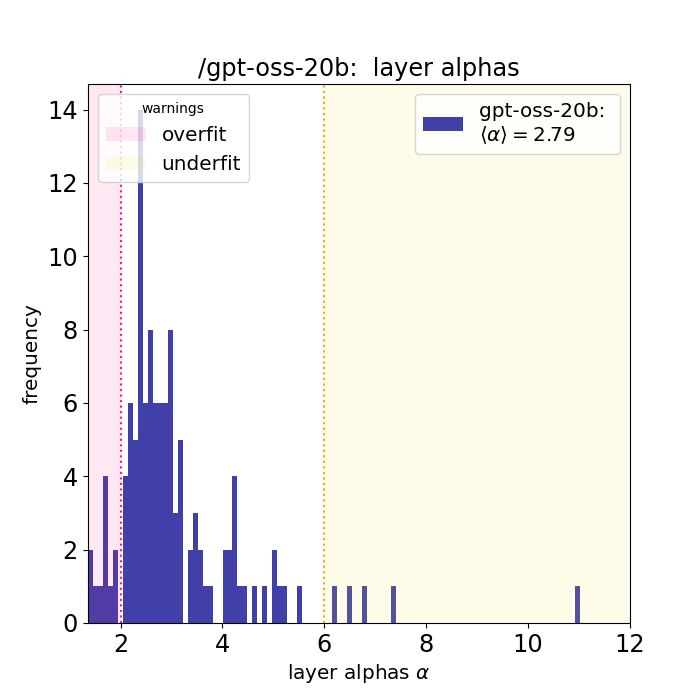}
  \end{minipage}\hfill
  \begin{minipage}[b]{0.4545\columnwidth}
    \centering
    \includegraphics[width=\linewidth]{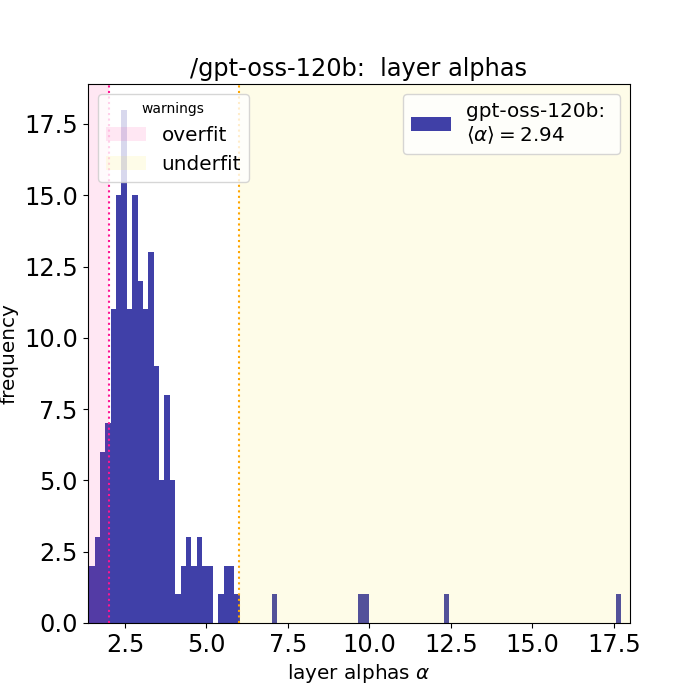}
  \end{minipage}
  \caption{Distribution of WeightWatcher HTSR PL $\alpha$ across layers of \texttt{gpt-oss-20b} and \texttt{gpt-oss-120b}. The dotted vertical lines mark the diagnostic warning thresholds: $\alpha \lesssim 2$ (pink) flags overfitting, while $\alpha \gtrsim 6$ (orange) flags underfitting. Both models exhibit a significant and unusual number of layers with $\alpha < 2$.}
  \label{fig:gptoss_layer_alpha_hist}
\end{figure}

These results underscore the utility of the open-source \texttt{WeightWatcher} tool for monitoring and understanding long-term generalization stability across different training schemes and models, with a particular strength in identifying overfitting and catastrophic collapse. For example, in Figure~\ref{fig:gptoss_layer_alpha_hist}, it is observed that production quality LLMs, like the OpenAI OSS GPT20B and 120B models show an unusually large number of layers with $\alpha<2$ (and Correlation Traps, not shown).  These results suggest these anomalies may hurt LLM performance in unpredictable ways.

%\FloatBarrier

\vspace{-0.5cm}

\section{Limitations}
\label{sec:limitations}

%Our study, while providing insights into generalization dynamics via Heavy-Tailed Self-Regularization (HTSR), has limitations that define important avenues for future research. The empirical findings are primarily derived from a specific three-layer MLP architecture trained on an MNIST subset. Consequently, the generalizability of the observed $\alpha$ trajectories and their specific predictive power for phenomena like grokking and late-stage generalization collapse warrants further validation across a wider range of model architectures (e.g., CNNs, Transformers), datasets, tasks, and diverse training configurations, including different optimizers and hyperparameter settings.

The \texttt{WeightWatcher} HTSR $\alpha$ metric is an empirically-grounded, phenomenological framework, supported rigorously with recently developed SETOL statistical mechanics theory. Still, care must be taken in interpreting the results with $\alpha$. While well-performing models exhibit $\alpha$ values within the range (e.g., $2 \le \alpha \le 6$), and $\alpha \approx 2$ is frequently associated with optimal performance, it is not guaranteed. It is conceivable that models might exhibit $\alpha$ values near or even below $2$ (typically indicating over-correlation) yet display suboptimal generalization. As shown here.  Other very-well trained models may have layers fairly large $\alpha\gg 2$, as is known.  This is not yet fully understood.  This highlights that while $\alpha$ provides strong correlational insights into learning phases and stability, the precise mapping of specific $\alpha$ values to absolute performance levels can be context-dependent and is an area for ongoing refinement of the theory. Our work contributes observations within specific phenomena, acknowledging that the broader applicability and predictive nuances when using the \texttt{WeightWatcher} tool and the HTSR and SETOL theories will benefit from continued exploration.
\bibliographystyle{icml2026}
\bibliography{references} % Assuming your references are in references.bib

@misc{power2022grokking,
  title         = {Grokking: Generalization Beyond Overfitting on Small Algorithmic Datasets},
  author        = {Alethea Power and Yuri Burda and Harri Edwards and Igor Babuschkin and Vedant Misra},
  year          = {2022},
  eprint        = {2201.02177},
  archivePrefix = {arXiv},
  primaryClass  = {cs.LG},
  url           = {https://arxiv.org/abs/2201.02177}
}

@misc{humayun2024deep,
  title         = {Deep Networks Always Grok and Here is Why},
  author        = {Ahmed Imtiaz Humayun and Randall Balestriero and Richard Baraniuk},
  year          = {2024},
  eprint        = {2402.15555},
  archivePrefix = {arXiv},
  primaryClass  = {cs.LG},
  url           = {https://arxiv.org/abs/2402.15555}
}

@inproceedings{liu2023omnigrokgrokkingalgorithmicdata,
  title         = {Towards understanding grokking: An effective theory of representation learning},
  author        = {Ziming Liu and Ouail Kitouni and Niklas S. Nolte and Eric J. Michaud and Max Tegmark and Mike Williams},
  booktitle     = {Advances in Neural Information Processing Systems},
  volume        = {35},
  pages         = {34651--34663},
  year          = {2022},
  editor        = {Surbhi Koyejo and Sham Kakade (formerly Mohamed) and Aarti Agarwal and Danielle Belgrave and Kyunghyun Cho and Alice Oh},
  publisher     = {Curran Associates, Inc.}
}

@misc{nanda2023progress,
  title         = {Progress measures for grokking via mechanistic interpretability},
  author        = {Neel Nanda and Lawrence Chan and Tom Lieberum and Jess Smith and Jacob Steinhardt},
  year          = {2023},
  eprint        = {2301.05217},
  archivePrefix = {arXiv},
  primaryClass  = {cs.LG},
  url           = {https://arxiv.org/abs/2301.05217}
}

@misc{varma2023explaining,
  title         = {Explaining grokking through circuit efficiency},
  author        = {Vikrant Varma and Rohin Shah and Zachary Kenton and J{\'a}nos Kram{\'a}r and Ramana Kumar},
  year          = {2023},
  eprint        = {2309.02390},
  archivePrefix = {arXiv},
  primaryClass  = {cs.LG},
  url           = {https://arxiv.org/abs/2309.02390}
}

@misc{golechha2024progressmeasuresgrokkingrealworld,
  title         = {Progress Measures for Grokking on Real-world Tasks},
  author        = {Satvik Golechha},
  year          = {2024},
  eprint        = {2402.11906},
  archivePrefix = {arXiv},
  primaryClass  = {cs.LG},
  url           = {https://arxiv.org/abs/2405.12755}
}

@misc{merrill2023tale,
  title         = {A tale of two circuits: grokking as competition of sparse and dense subnetworks},
  author        = {William Merrill and Nikolaos Tsilivis and Aman Shukla},
  year          = {2023},
  eprint        = {2303.11873},
  archivePrefix = {arXiv},
  primaryClass  = {cs.LG},
  url           = {https://arxiv.org/abs/2303.11873}
}

@misc{peng2023theoretical,
  title         = {Theoretical Explanation of Activation Sparsity through Flat Minima and Adversarial Robustness},
  author        = {Ze Peng and Lei Qi and Yinghuan Shi and Yang Gao},
  year          = {2023},
  eprint        = {2309.03004},
  archivePrefix = {arXiv},
  primaryClass  = {cs.LG},
  url           = {https://arxiv.org/abs/2309.03004}
}

@article{martin2021implicit,
  author    = {Martin, Charles H. and Mahoney, Michael W.},
  title     = {Implicit Self-Regularization in Deep Neural Networks: Evidence from Random Matrix Theory and Implications for Learning},
  journal   = {Journal of Machine Learning Research},
  volume    = {22},
  number    = {1},
  pages     = {1--73},
  year      = {2021},
  month     = jan,
  eid       = {165},
  url       = {http://jmlr.org/papers/v22/20-410.html},
}

@article{martin2021predicting,
  author  = {Martin, Charles H. and Peng, Tian and Mahoney, Michael W.},
  title   = {Predicting trends in the quality of state-of-the-art neural networks without access to training or testing data},
  journal = {Nature Communications},
  volume  = {12},
  pages   = {4122},
  year    = {2021},
  month   = {jul},
  doi     = {10.1038/s41467-021-24025-8},
  url     = {https://doi.org/10.1038/s41467-021-24025-8},
}

@article{marchenko1967distribution,
  title         = {Distribution of eigenvalues for some sets of random matrices},
  author        = {Vladimir A. Marchenko and Leonid Andreevich Pastur},
  journal       = {Matematicheskii Sbornik},
  volume        = {72(114)},
  number        = {4},
  pages         = {507--536},
  year          = {1967},
  publisher     = {Steklov Mathematical Institute of Russian Academy of Sciences}
}

@article{clauset2009power,
  title         = {Power-Law Distributions in Empirical Data},
  author        = {Aaron Clauset and Cosma Rohilla Shalizi and Mark E.J. Newman},
  journal       = {SIAM Review},
  volume        = {51},
  number        = {4},
  pages         = {661--703},
  year          = {2009},
  publisher     = {SIAM}
}

@misc{weightwatcher,
  title         = {{WeightWatcher}: Analyze {Deep Learning} {Models} without {Training} or {Data}},
  author        = {Charles H. Martin},
  year          = {2018-2024},
  howpublished  = {\url{https://github.com/CalculatedContent/WeightWatcher}},
  note          = {Version 0.7.5.5 used in this study. Accessed May 12, 2025}
}

@misc{huesmann2021impact,
  title         = {The Impact of Activation Sparsity on Overfitting in Convolutional Neural Networks},
  author        = {Karim Huesmann and Luis Garcia Rodriguez and Lars Linsen and Benjamin Risse},
  year          = {2021},
  eprint        = {2104.06153},
  archivePrefix = {arXiv},
  primaryClass  = {cs.LG},
  url           = {https://arxiv.org/abs/2104.06153},
}

@inproceedings{li2023lazyneuronphenomenonemergence,
  title         = {The Lazy Neuron Phenomenon: On Emergence of Activation Sparsity in Transformers},
  author        = {Zonglin Li and Chong You and Srinadh Bhojanapalli and Daliang Li and Ankit Singh Rawat and Sashank J. Reddi and Ke Ye and Felix Chern and Felix Yu and Ruiqi Guo and Sanjiv Kumar},
  booktitle     = {The Eleventh International Conference on Learning Representations (ICLR)},
  year          = {2023},
  url           = {https://openreview.net/forum?id=TJ2nxciYCk-},
  note          = {arXiv:2210.06313}
}

@article{martin2025setol,
title = {{SETOL}: A Semi-Empirical Theory of (Deep) Learning},
  author  = {Martin, Charles H. and Hinrichs, Christopher},
  journal = {arXiv preprint arXiv:2507.17912},
  year    = {2025},
  url     = {https://arxiv.org/abs/2507.17912}
}

@article{prakash2025grokking,
  title   = {Grokking and Generalization Collapse: Insights from HTSR theory},
  author  = {Prakash, Hari K. and Martin, Charles H.},
  journal = {arXiv preprint},
  volume  = {arXiv:2506.04434},
  year    = {2025},
  url     = {https://arxiv.org/abs/2506.04434}
}

% --- Appendix Starts Here ---
%\newpage % Optional: start appendix on a new page
% --- Appendix Starts Here ---
\newpage % Optional: start appendix on a new page
\onecolumn
\begin{appendices} % Use the appendices environment

% Original Appendix A (Experimental Setup) moved to main paper as Section 4.
% Original Appendix B (Experiment with Weight Decay) is now Appendix A.
% Original Appendix C (Examples of Heavy-Tailed Spectral Fits) is now Appendix B, with fig:pl_fit_example and its description moved to main paper.

\section{MLP Experimental Setup}
\label{app:exp_setup} % Changed label from app:setup to app:exp_setup for consistency

We train a Multi-Layer Perceptron (MLP) on a subset of the MNIST dataset using the hyperparameters detailed in Table~\ref{tab:hyperparameters_appendix}. The training subset is constructed by randomly selecting 100 samples from each of the 10 MNIST classes, ensuring a balanced dataset of 1,000 unique training points.
This was run on an Nvidia Quadro P2000 and took approximately 11 hours. A considerable part of the time is due to the speed of saving the measures.

\begin{table}[h!]
\centering
\caption{MLP Experimental hyperparameters used in the study (details in Appendix~\ref{app:exp_setup}).} % Modified caption
\vspace{0.2cm}
\label{tab:hyperparameters_appendix} % Changed label from tab:hyperparameters
\begin{tabular}{@{}ll@{}}
\toprule
\textbf{Parameter}        & \textbf{Value}                                                              \\
\midrule
Network Architecture    & Fully Connected MLP                                                         \\
Depth                   & 3 Linear layers (Input $\to$ Hidden1 $\to$ Hidden2 $\to$ Output)                \\
Width                   & 200 hidden units per hidden layer                                           \\
Activation Function     & ReLU (Rectified Linear Unit)                                                \\
Input Layer Size        & 784 (Flattened MNIST image $28 \times 28$)                                    \\
Output Layer Size       & 10 (MNIST digits 0-9)                                                       \\
Weight Initialization   & Default PyTorch (Kaiming Uniform for weights), parameters scaled by 8.0 \\
Bias Initialization     & Default PyTorch (Uniform), then scaled by 8.0                               \\
Dataset                 & MNIST                                                                       \\
Training Points         & 1,000 (100 per class, stratified random sampling)                           \\
Test Points             & Standard MNIST test set (10,000 samples)                                    \\
Batch Size              & 200                                                                         \\
Loss Function           & Mean Squared Error (MSE) with one-hot encoded targets                       \\
Optimizer               & AdamW                                                                       \\
Learning Rate (LR)      & $5 \times 10^{-4}$                                                            \\
Weight Decay (WD)       & 0.0 (for main results), 0.01 (for Appendix~\ref{app:wd_experiment} comparison) \\
AdamW $\beta_1$         & 0.9 (PyTorch default)                                                       \\
AdamW $\beta_2$         & 0.999 (PyTorch default)                                                     \\
AdamW $\epsilon$        & $10^{-8}$ (PyTorch default)                                                   \\
Optimization Steps      & $10^7$                                                                      \\
Data Type (PyTorch)     & `torch.float64`                                                             \\
Random Seed             & 0 (for all libraries)                                                       \\
Software Framework      & PyTorch                                                                     \\
HTSR Tool               & WeightWatcher v0.7.5.5 \cite{weightwatcher}                                 \\
\bottomrule
\end{tabular}
\end{table}

\bigskip
\noindent \textbf{Note on Weight Decay:} The primary results presented in this paper, particularly those demonstrating grokking followed by late-stage generalization collapse (Figure~\ref{fig:training_curves}), were obtained with weight decay explicitly set to 0. This allows observation of the learning dynamics driven purely by the optimizer and the loss landscape while exhibiting both phenomena, whereas the other proposed measures fail to detect the grokking transition of increasing test accuracy. Runs with non-zero weight decay (e.g., \texttt{WD=0.01}, see Appendix~\ref{app:wd_experiment}) were also performed for comparison, showing different dynamics but confirming the general utility of HTSR.

\section{Comparative Grokking Progress Metrics and Measures}
\label{app:comparative_measures}

\paragraph{Weight Norm Analysis}
Following observations that weight decay can influence grokking \cite{liu2023omnigrokgrokkingalgorithmicdata}, we monitor the $\ell_2$ norm of the network's weights, 
\begin{equation}
    ||\mathbf{W}||_2 = \sqrt{\sum_{l} ||\mathbf{W}_l||_F^2},
\end{equation}
throughout training. We specifically run experiments with weight decay disabled (\texttt{WD=0}) to isolate the effect of the optimization dynamics on the norm itself. 
%As shown in Figure~\ref{fig:training_curves}, grokking still occurs robustly even when the weight norm steadily increases, indicating that while related, the $\ell_2$ norm alone is not the determining factor for the grokking transition or its subsequent stability.

%We also compute the progress measures proposed by Golechha et al.~\cite{golechha2024progressmeasuresgrokkingrealworld} that builds on prior work to capture broader network changes.

\paragraph{Activation Sparsity.}
For a given layer with activations $b_{i,j}$ (representing the activation of neuron $j$ for input example $i$), the activation sparsity $\Lambda_{AS}$ is defined as:
\begin{equation}
A_s = \frac{1}{T}\sum_{i=1}^{T} \frac{1}{n}\sum_{j=1}^{n}\mathbf{1}(b_{i,j} < \tau),
\end{equation}
where $T$ is the number of training examples, $n$ is the number of neurons in the layer, $\tau$ is a chosen threshold, and $\mathbf{1}(\cdot)$ is the indicator function. This metric measures neuron inactivity. Prior studies have linked activation sparsity to generalization \cite{li2023lazyneuronphenomenonemergence, merrill2023tale, peng2023theoretical} and reported specific dynamics such as plateauing before grokking \cite{golechha2024progressmeasuresgrokkingrealworld} or an increase preceding a rise in test loss \cite{huesmann2021impact}. 
%In our primary WD=0 experiments, $\Lambda_{AS}$ generally increases throughout training (Figure~\ref{fig:other_measures}). We observe a subtle inflection or dip in $\Lambda_{AS}$ coinciding with the point of maximum test accuracy before a slight increase. This feature's predictive utility would need to be further researched, as this dip does not completely indicate the magnitude of decay that follows. (e.g., WD=0.01, see Appendix~\ref{app:wd_experiment}).

\paragraph{Absolute Weight Entropy.}
For a weight matrix $W \in \mathbb{R}^{m \times n}$, the absolute weight entropy $H_{abs}(W)$ is given by:
\begin{equation}
H_{abs}(W) = -\sum_{i=1}^{m}\sum_{j=1}^{n}|w_{i,j}|\log|w_{i,j}|.
\end{equation}
This entropy quantifies the spread of absolute weight magnitudes. Golechha ~\cite{golechha2024progressmeasuresgrokkingrealworld} suggested its sharp decrease signals generalization.
%We observe however, $H_{abs}(W)$ also decreases sharply during the late-stage generalization collapse (Figure~\ref{fig:other_measures}). 

\paragraph{Approximate Local Circuit Complexity.}
Let $L^{(W)}(x)$ denote the output logits for input $x$ using weights $W$, and let $L^{(W')}(x)$ denote the logits when 10\% of the weights are set to zero (forming $W'$). The approximate local circuit complexity, denoted $\Lambda_{LC}$, is the summed KL divergence:
\begin{equation}
\Lambda_{LC} = \sum_{k=1}^{N_{data}}\sum_{j \in \mathcal{C}} \text{Pr}\bigl(j | L^{(W)}(x_k)\bigr) \log\frac{\text{Pr}\bigl(j | L^{(W)}(x_k)\bigr)}{\text{Pr}\bigl(j | L^{(W')}(x_k)\bigr)}.
\end{equation}
Here, $N_{data}$ is the number of training examples $x_k$, $\mathcal{C}$ is the set of classes, and $\text{Pr}(j | L(x))$ is the probability of class $j$ derived from the logits $L(x)$ (e.g., via softmax). This measure captures output sensitivity to minor weight perturbations. Lower $\Lambda_{LC}$ has been linked to stable, generalizable representations \cite{golechha2024progressmeasuresgrokkingrealworld}.
%we find it remains low throughout the late-stage generalization collapse (Figure~\ref{fig:other_measures}). This challenges its utility as a consistent indicator of robust generalization, as it fails to reflect this performance catastrophe.

\section{MLP Experiment without Weight Decay} % Was Appendix B
\begin{figure*}[!ht]
  \centering
  \includegraphics[width=0.95\textwidth]{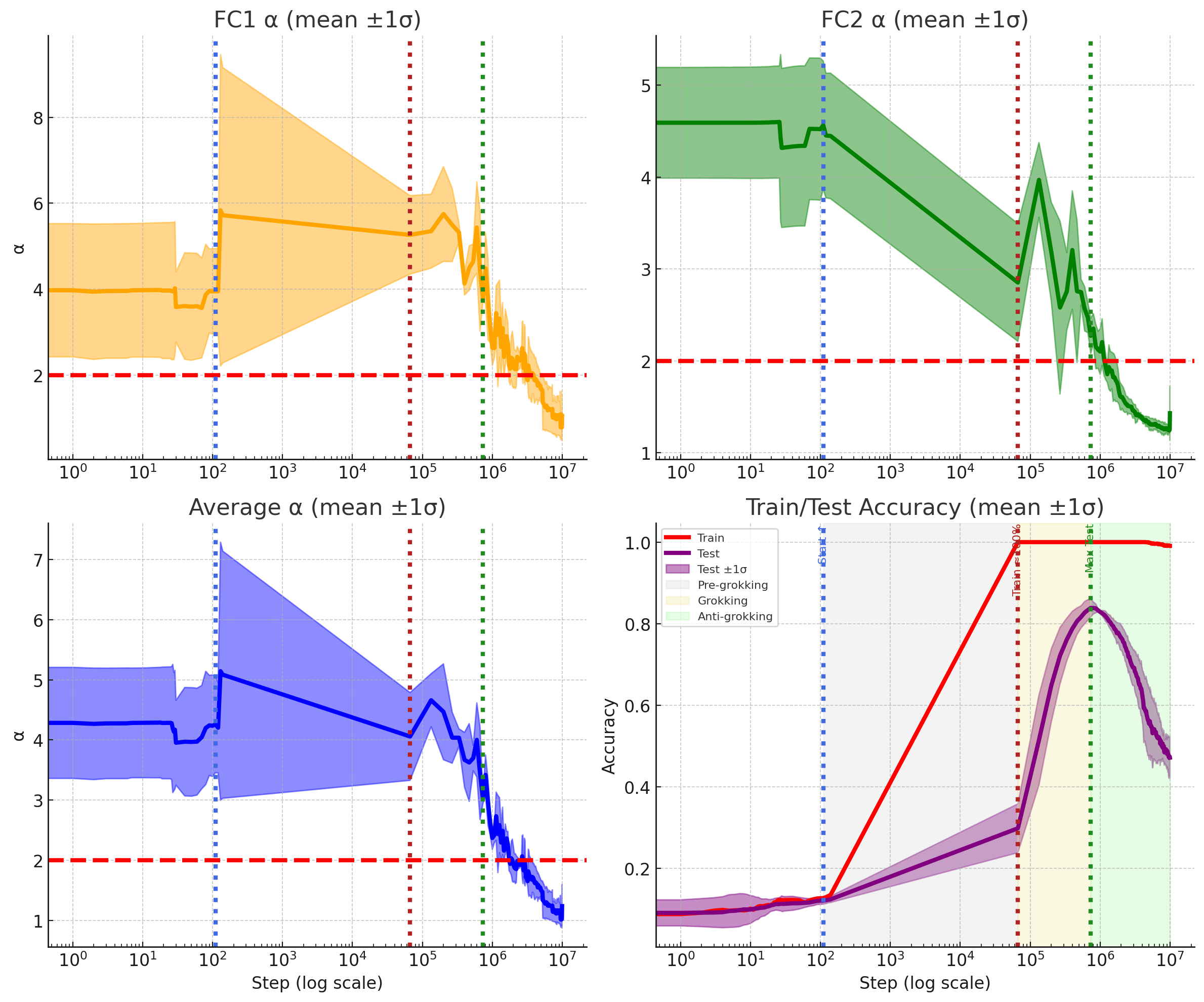}
  \caption{\textbf{Layer-wise $\alpha$ dynamics and accuracy across training (replicate run).}
  \textbf{Top left}: Mean $\alpha$ (±1$\sigma$) for the first fully connected layer (FC1).
  \textbf{Top right}: Mean $\alpha$ (±1$\sigma$) for the second fully connected layer (FC2).
  \textbf{Bottom left}: Mean $\alpha$ averaged across layers.
  \textbf{Bottom right}: Train and test accuracy (mean ±1$\sigma$).
  Vertical dashed lines indicate the transitions between pre-grokking (blue), grokking (red), and anti-grokking (green) phases.
  The dashed horizontal line at $\alpha=2$ marks the HTSR/SETOL critical value.
  During grokking, layer-wise $\alpha$ approaches the optimal regime $\alpha \approx 2$, coinciding with peak test accuracy.
  In the anti-grokking phase, $\alpha$ drops below 2 in one or more layers, coinciding with generalization collapse.}
  \label{fig:replicate-alpha}
\end{figure*}

Figure~\ref{fig:replicate-alpha} provides the full HTSR layer $\alpha$ for the \texttt{WD=0} experiment, for each layer, plotted for all three phases.

\section{MLP Experiment with Weight Decay} % Was Appendix B
\label{app:wd_experiment} % Label remains the same, cross-refs should still work

\begin{figure}[h!]
    \centering
    \includegraphics[width=0.8\textwidth]{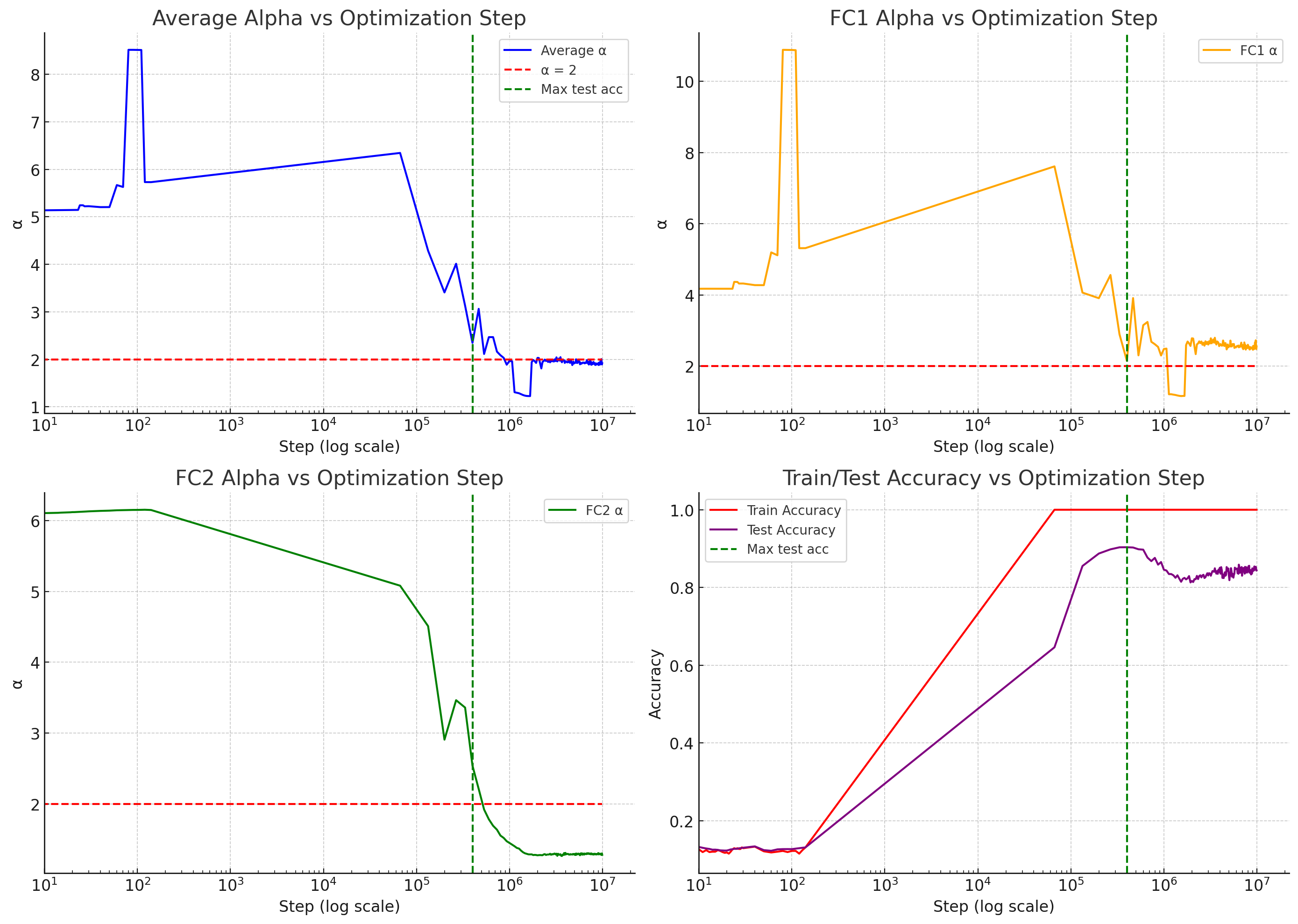}
    \caption{HTSR $\alpha$ exponent evolution for the MLP trained with WD=0.01.}
    \label{fig:appendix_alpha_wd}
\end{figure}

\begin{figure}[h!]
    \centering
    \includegraphics[width=0.8\textwidth]{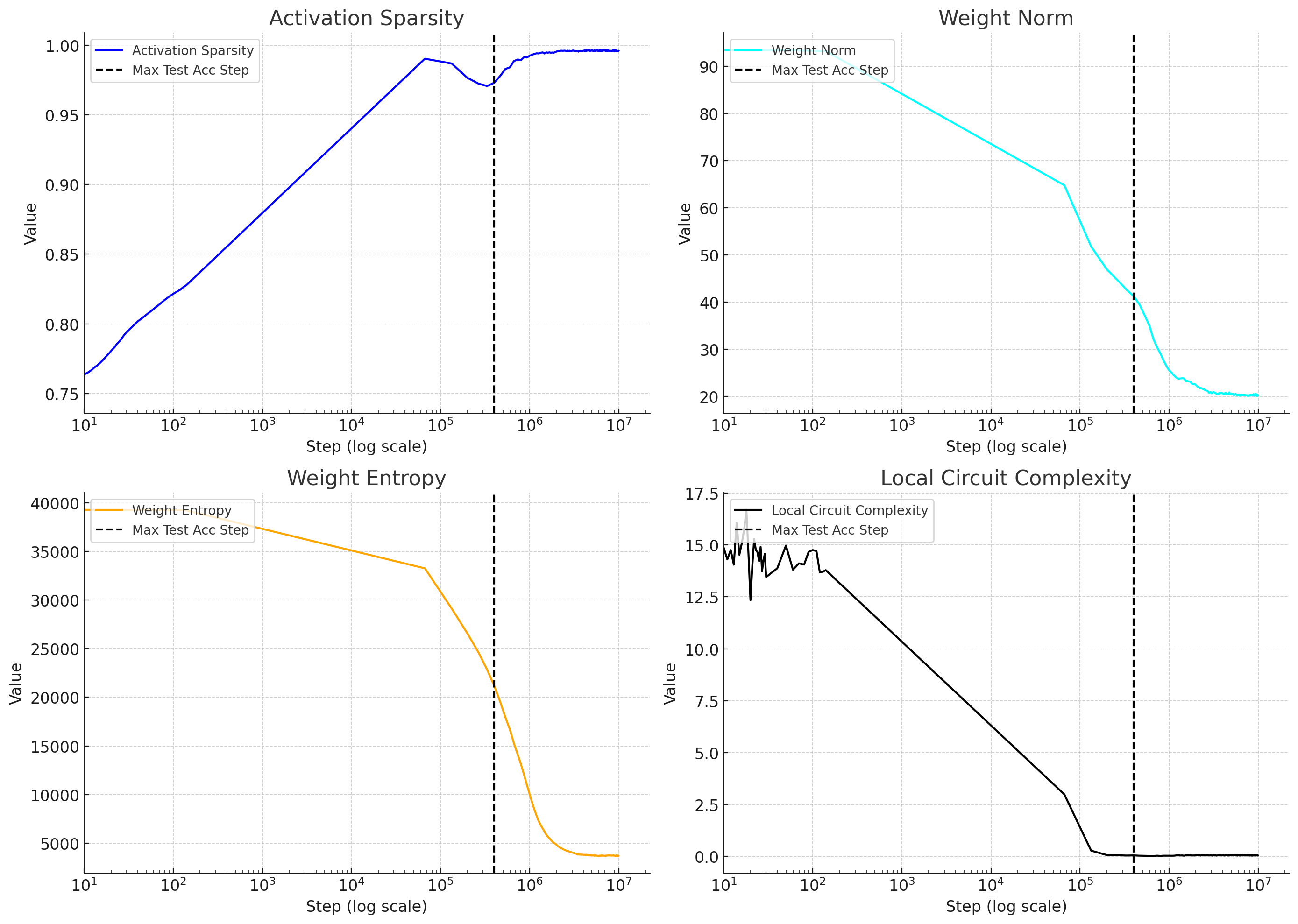}
    \caption{Progress measures (Activation Sparsity, Weight Entropy, Circuit Complexity) and $\ell_2$ Weight Norm for the MLP trained with WD=0.01.}
    \label{fig:appendix_other_measures_wd}
\end{figure}

To further understand the influence of weight decay on the observed generalization dynamics and the behavior of our tracked metrics, we conducted an experiment identical to our main study (\texttt{WD=0}) but with a small amount of weight decay (\texttt{WD=0.01}) applied. The training curves and metric evolutions for this \texttt{WD=0.01} experiment are presented in Figures
\ref{fig:appendix_alpha_wd}, and \ref{fig:appendix_other_measures_wd}.

A key characteristic of training with weight decay is the tendency for the $\ell_2$ norm of the weights to decrease over time, or stabilize at a lower value, which is observed in this experiment (Figure~\ref{fig:appendix_other_measures_wd}). This contrasts with the continuously increasing  $\ell_2$ weight norm seen in our primary \texttt{WD=0} experiments.

In this WD=0.01 regime, the network still achieves a high level of test accuracy. Notably, after the initial grokking phase, the test accuracy slightly decreases and then enters a prolonged plateau, maintaining near peak performance for a significant number of optimization steps (Figure~\ref{fig:appendix_alpha_wd}). Correspondingly, the average heavy-tail exponent, $\alpha$, also exhibits the decrease and a distinct plateau around the critical value of $\alpha \approx 2$ during this period (Figure~\ref{fig:appendix_alpha_wd}, top left panel). 

The other progress measures considered—Activation Sparsity and Approximate Local Circuit Complexity—also tend to plateau or stabilize during this phase of peak test performance in the \texttt{WD=0.01} setting (Figure~\ref{fig:appendix_other_measures_wd}).
This contrasts with the \texttt{WD=0} scenario where, despite eventual grokking, the system does not find such a stable long-term plateau and instead proceeds towards a late-stage generalization collapse. The observation that $\alpha$ (and other metrics) plateau in conjunction with peak, stable test accuracy under traditional weight decay settings aligns with some existing understanding of well-regularized training.

While HTSR and the $\alpha$ exponent provide valuable insights in both regimes, its unique capability to signal impending collapse in the absence of weight decay underscores its importance for understanding layer dynamics under various scenarios.

% \begin{figure}[h!]
%     \centering
%     % \includegraphics[width=0.8\textwidth]{path_to_your_appendix_training_curves_wd_figure.png}
%     \caption{Training and test accuracy/loss for the MLP trained with WD=0.001.}
%     \label{fig:appendix_training_curves_wd}
% \end{figure}

\FloatBarrier
\FloatBarrier

\section{Statistical Analysis and Validation of Correlation Traps}
\label{app:stats_of_traps}

Here, to further validate the presence of Correlation Traps for the zero weight decay \texttt{WD=0} experiment , we report the results of statistical tests designed to determine if the randomized ESD of the $\mathbf{W}^{rand}$ fits an MP distribution or not.  Briefly we fit the ESD to a MP distribution and report the fitted variance $\sigma_{mp}$, the Kolmogorov-Smirnov (KS) statistic of the fit, and the p-value for the MP fit as the null model. 
We also report the number of Correlation Traps, as determined using the open-source \texttt{WeightWatcher} tool\cite{weightwatcher}.
Results for layer FC1 are presented in Table~\ref{tab:ks_test_traps_no_layer}.   
Results for FC2 are similar (not shown).
 Additional details are provided in the supplementary material.

\begin{table}[h!]
\centering
\caption{\textbf{Statistical validation of Correlation Traps.} Selected results for layer FC1 at different training stages for zero weight decay (\texttt{WD=0}) experiment. MP Variance $(\sigma_{MP})$ Kolmogorov-Smirnov (KS) test statistic, p-value for MP fit, and  number of detected Correlation Traps.   
Pre-grokking $\sim10^5$ steps, 
 Grokking $10^6$ steps, and Anti-grokking $10^7$ steps,
}
\vspace{0.2cm}
\label{tab:ks_test_traps_no_layer}
\begin{tabular}{@{}lcccc@{}}
\toprule
\textbf{Model State} & \textbf{MP variance} $(\sigma_{mp})$ & \textbf{KS Statistic} & \textbf{p-value} & \textbf{\# Traps} \\
\midrule
Pre-Grokking &$\approx 1.002$ & 0.0120 & $\approx 1.0$ & 0 \\
Grokking (Max Test Accuracy) &$\approx 0.999$ & 0.0212 & $\approx 1.0$ & 0 \\
Anti-Grokking (Collapse) &$\approx 0.949$ & 0.3044 & $1.877 \times 10^{-5}$ & 9 \\
\bottomrule
\end{tabular}
\end{table}

\paragraph{Initial Layer State (Pre-Grokking \texttt{WD=0}):}
Immediately after initialization, the network weights are expected to be largely random, and their ESD should conform well to the MP distribution. Figure~\ref{fig:ESDs} (Right) shows an MP fit to an ESD from a representative layer $\mathbf{W}^{rand}$ of the newly initialized model. A KS test comparing this empirical ESD to the fitted MP distribution (using $\sigma_{mp} \approx 1.0024$ as estimated by \texttt{WeightWatcher}) yielded a KS statistic of $0.0120$ and a p-value $\approx 1.0$. This high p-value indicates this ESD is statistically consistent with the MP distribution, as expected.

\paragraph{Best Layer State (Grokking phase \texttt{WD=0}):}
As the network learns and reaches its maximum test accuracy, significant structure develops in the elements of the weight matrices $W_{i,j}$. This can be seen by randomizing the layer weight matrix elementwise, $\mathbf{W}\rightarrow\mathbf{W}^{rand}$ , and  plotting ESD, and looking for deviations from the theoretical MP distribution. The ESD now typically exhibits a pronounced heavy tail, with eigenvalues extending beyond the bulk region that might be approximated by an MP fit. For our model at peak test accuracy, the KS test against a fitted MP model ($\sigma_{mp} \approx 0.999$) resulted in a KS statistic of $0.0212$ and a p-value $\approx 1$.  Again, we cannot reject the MP null model.

%This extremely small p-value signifies a very strong statistically significant deviation from the MP null model, highlighting the emergence of the Correlation Traps.

\paragraph{Final Layer State (Anti-Grokking phase \texttt{WD=0}):}
In the late-stage of training, as the model undergoes generalization collapse and enters an over-correlated state (characterized by $\alpha < 2$), the ESD of $\mathbf{W}^{rand}$ structure continues to reflect a non-random configuration. The KS test for the final model against an MP fit (with an estimated $\sigma_{mp} \approx 2$) yielded a KS statistic of $0.3044$ and a p-value of $1.877 \times 10^{-5}$ Figure~\ref{fig:Traps} (Right).  This result, when interpreted jointly with the evidence of the large outliers, further confirms that the network's structure remains significantly different from a random matrix baseline, consistent with the highly correlated or near rank-collapsed state indicated by our HTSR analysis.

% \textbf{THIS BELONGS SOMEWERE HERE}: the FC2 layer $\alpha$ decreased significantly $(\alpha < 3)$, whereas the FC1 layer $\alpha$ remains quite large  $(\alpha > 5.5)$.  (Note that layer FC3 is too small to analyze).  

These quantitative comparisons demonstrate a transition from an initially random-like state (consistent with MPD) to progressively more structured, non-random states as learning occurs and eventually leads to over-correlation. The inability of the MP distribution to describe these learned features, especially the heavy tails, necessitates the use of tools like the HTSR theory, the PL exponent $\alpha$, and the open-source \texttt{WeightWatcher} tool, to properly characterize these complex correlation structures and their relationship to generalization performance.

\section{Conditions for Outliers after Entry-wise Shuffling of a Weight Matrix}
\label{app:shuffle_outliers}

\paragraph{Neural-network layer setting.}
Consider a fully connected layer with weight matrix $W\in\mathbb{R}^{M\times N}$ mapping an $N$-dimensional input to an $M$-dimensional output, and aspect ratio
\[
\gamma \;:=\; \frac{M}{N}\in(0,\infty).
\]

\paragraph{Notation.}
For any matrix $W$ set
\[
X(W):=\frac{1}{N}\,W^{\!\top}W,
\qquad
\lambda_{\max}(W):=\lambda_{\max}\!\bigl(X(W)\bigr).
\]
Fix a variance proxy $\sigma^{2}>0$ and define the Marchenko–Pastur (MP) edges
\[
\lambda_{-}:=\sigma^{2}\bigl(1-\sqrt{\gamma}\bigr)^{2},
\qquad
\lambda_{+}:=\sigma^{2}\bigl(1+\sqrt{\gamma}\bigr)^{2}.
\]

\paragraph{Limit convention.}
As is standard for MP limits, let $N\to\infty$ with $M/N\to\gamma$; every limit below is taken along this sequence.

\paragraph{Norm conventions.}
For $x\in\mathbb{R}^{N}$, $\lVert x\rVert_{2}$ is the Euclidean norm; for $A\in\mathbb{R}^{M\times N}$, $\lVert A\rVert$ is the spectral norm; and for symmetric $S$, $\lambda_{\max}(S)$ is its top eigenvalue.

\paragraph{Marchenko–Pastur reminder.}
The $\mathrm{MP}_{\sigma^{2},\gamma}$ density is
\[
\rho_{\mathrm{MP}}(\lambda)
=\frac{\sqrt{(\lambda_{+}-\lambda)(\lambda-\lambda_{-})}}{2\pi\,\sigma^{2}\,\gamma\,\lambda}\;
\mathbf{1}_{[\lambda_{-},\lambda_{+}]}(\lambda).
\]

\paragraph{Element-wise shuffling (scrambling).}
Let $\pi:\{1,\dots,M\}\times\{1,\dots,N\}\!\to\!\{1,\dots,M\}\times\{1,\dots,N\}$ be any permutation. The scrambled matrix is
\[
W_{\mathrm{rand}}:=(W_{ij})_{\pi(i,j)}.
\]
Define $A_{N}:=X\!\bigl(W^{(N)}_{\mathrm{rand}}\bigr)$, so that
\[
\lambda_{\max}(A_{N})=\lambda_{\max}\!\bigl(X(W^{(N)}_{\mathrm{rand}})\bigr).
\]

\noindent\textbf{Definition 1 (typical vs.\ atypical).}
A sequence $\{W^{(N)}\}$ is \emph{typical} if $\lambda_{\max}(A_{N})\to\lambda_{+}$, and \emph{atypical} if
\[
\liminf_{N\to\infty}\!\bigl[\lambda_{\max}(A_{N})-\lambda_{+}\bigr]>0.
\]

\noindent\textbf{Theorem 1 (single-entry sufficient condition).}
Assume $M/N\to\gamma\in(0,\infty)$ and for some index pair $(p_{N},q_{N})$,
\[
\frac{|W^{(N)}_{p_{N}q_{N}}|}{\sqrt{N}}\;\xrightarrow[N\to\infty]{}\;\theta,
\qquad
\theta>\sigma\bigl(1+\sqrt{\gamma}\bigr).
\]
Then there exists $N_{0}$ such that for all $N\ge N_{0}$, $\lambda_{\max}(A_{N})>\lambda_{+}$, so the layer is \emph{atypical}.

\emph{Proof.}
Let $a_{N}:=\max_{i,j}|W_{ij}|$. Permuting entries does not change $a_{N}$. Choose a column in $W_{\mathrm{rand}}$ that contains an entry of magnitude $a_{N}$. Denote by $e_{c_{N}}\in\mathbb{R}^{N}$ the corresponding standard basis vector (with a $1$ in position $c_{N}$ and zeros elsewhere). Since $\lVert e_{c_{N}}\rVert_{2}=1$, the Rayleigh quotient yields
\[
e_{c_{N}}^{\!\top}A_{N}e_{c_{N}}
=\frac{1}{N}\,\lVert W_{\mathrm{rand}}e_{c_{N}}\rVert_{2}^{2}
=\frac{1}{N}\sum_{i=1}^{M} |(W_{\mathrm{rand}})_{i c_{N}}|^{2}
\;\ge\; \frac{a_{N}^{2}}{N}.
\]
By the Courant–Fischer principle, for any symmetric matrix $S$,
\[
\lambda_{\max}(S)
  = \max_{\lVert x\rVert_2 = 1} x^{\!\top} S x,
\]
hence $\lambda_{\max}(A_{N}) \ge a_{N}^{2}/N$. By the hypothesis,
\[
a_{N}\;\ge\; \bigl|W^{(N)}_{p_{N}q_{N}}\bigr|
             \;=\;(\theta+o(1))\sqrt{N},
\]
so that $a_{N}^{2}/N \to \theta^{2}$. Because $\theta > \sigma(1+\sqrt{\gamma})$, we have
\[
\theta^{2} \;>\; \sigma^{2}(1+\sqrt{\gamma})^{2} \;=\; \lambda_{+}.
\]
Consequently, there exists $N_{0}$ such that for all $N\ge N_{0}$,
\[
\lambda_{\max}(A_{N}) \;\ge\; \frac{a_{N}^{2}}{N} \;>\; \lambda_{+}.
\]
Therefore the layer under this condition is \emph{atypical}. 

\paragraph{Connection with the BBP transition.}
In the classic spiked-covariance model $W=Z+\theta\,uv^{\top}$ (rank-one signal plus i.i.d.\ noise), an eigenvalue separates from the Marchenko–Pastur bulk precisely when
\[
\theta^{2}=\sigma^{2}(1+\sqrt{\gamma})^{2}.
\]
The condition above places the scrambled matrix in the super-critical regime
\[
\theta^{2}>\sigma^{2}(1+\sqrt{\gamma})^{2},
\]
so an outlier survives even after correlations are destroyed.

\paragraph{Discussion and intuition.}
Modern empirical work suggests that \emph{stochastic gradient descent} (SGD) can drive a handful of coordinates of a weight matrix toward abnormally large magnitudes, producing \emph{elementwise} heavy-tailed weight distributions[2]. Here, these heavy-tailed elements $W_{i,j}$ appear as rank-one perturbations, or \emph{spikes}, that distort the spectrum of the layer’s Gram matrix; eigenvalues above the MP bulk appear precisely when the test accuracy drops catastrophically.

Entry-wise shuffling deliberately \emph{destroys all remaining correlations} in the weight matrix while preserving the multiset of magnitudes. If a large eigenvalue survives the shuffle, there must be a \emph{pure-magnitude} signal, rendering the layer atypical.

Theorem~1 and BBP shows that the presence of even one entry growing such that
\[
\theta^{2} \;>\; \sigma^{2}\bigl(1+\sqrt{\gamma}\bigr)^{2},
\]
suffices to guarantee such a surviving outlier, confirming the empirical link between unusually large coordinates and atypical spectra.

This is also highlighted in Reference~[3], see Section~5.3 (Bulk\,+\,Spikes).

\paragraph{References.}
\begin{enumerate}\setlength{\itemsep}{0.2em}
\item Jinho Baik, G\'erard Ben Arous, and Sandrine P\'ech\'e. \emph{Phase transition of the largest eigenvalue for nonnull complex sample covariance matrices}. The Annals of Probability, Ann.\ Probab.\ 33(5), 1643--1697 (September 2005).
\item M. Gurbuzbalaban, U. Simsekli, and L. Zhu. \emph{The heavy-tail phenomenon in SGD}. In \emph{International Conference on Machine Learning} (ICML), pp.\ 3964--3975. PMLR, 2021.
\item Charles H. Martin and Michael W. Mahoney. \emph{Implicit self-regularization in deep neural networks: Evidence from random matrix theory and implications for learning}. Journal of Machine Learning Research, 22(165):1--73, 2021.
\item Charles H. Martin and Christopher Hinrichs. \emph{SETOL: A Semi-Empirical Theory of (Deep) Learning}. arXiv preprint arXiv:2507.17912, 2025.
\end{enumerate}

% start of section on eigenvector analysis

\section{Largest Tail Eigenvectors and Correlation Traps}
\label{sec:tail_vectors}

We now analyze the \emph{largest eigenvalues/eigenvectors} of layer Gram matrices
and relate them to the \emph{Correlation Traps} detected in the anti-grokking
phase. Throughout, for a weight matrix $W\!\in\!\mathbb{R}^{M\times N}$ we write
\[
X(W)\;:=\;\frac{1}{N}\,W^{\!\top}W,
\qquad
\lambda_{\max}(W)\;:=\;\lambda_{\max}\!\bigl(X(W)\bigr),
\]
and recall the Marchenko–Pastur (MP) upper edge
$\lambda_{+}=\sigma^{2}(1+\sqrt{\gamma})^{2}$ with $\gamma=M/N$.

We call any eigenpair $(\lambda,v)$ of $X(W)$ as defined by the start of the power law as 
a \emph{tail} eigenpair, and when $\lambda>\lambda_{+}$ for the
\emph{scrambled} matrix $W_{\mathrm{rand}}$ (entry-wise permutation of $W$;
see App.~\ref{app:shuffle_outliers}) we call $(\lambda,v)$ a
\emph{trap} eigenpair. These survive after all inter-entry correlations have been
destroyed by shuffling.

\paragraph{What could cause traps? }
App.~\ref{app:shuffle_outliers} proves a simple sufficient condition:
if even a single entry satisfies
$\lvert W_{p_N q_N}\rvert/\sqrt{N}\!\to\!\theta$ with
$\theta>\sigma(1+\sqrt{\gamma})$, then for all large $N$,
$\lambda_{\max}\!\bigl(X(W_{\mathrm{rand}})\bigr)>\lambda_{+}$.
Thus one cause can be \emph{rank-one perturbations in the weight matrix} that create an MP-outlying
eigenvalue that \emph{survives} shuffling, i.e., a trap. This is the
BBP super-critical regime and constitutes a purely \emph{magnitude-driven}
explanation for atypical spectra, independent of correlations.

\paragraph{Empirical evidence for the Structural outlier.}
Weight-value histograms across checkpoints (initial $\rightarrow$ best-test
$\rightarrow$ anti-grok) exhibit these structural outliers. For a 3-layer MLP trained on MNIST:
\begin{itemize}
\item \textbf{First layer ($W_1\!\in\!\mathbb{R}^{200\times 784}$).}
  Anti-grokking shows extreme coordinates (e.g., $\approx-14.78$ and $\approx+5.67$),
  well outside the best-test and pre-grok ranges; the bulk remains
  near zero but with markedly heavier tails.
\item \textbf{Second layer ($W_2\!\in\!\mathbb{R}^{200\times 200}$).}
  Anti-grokking again introduces large coordinates (e.g., $\approx-4.68$, $\approx+1.11$)
  absent at pre-grok and subdued at best-test(grok).
\item \textbf{Third layer ($W_3\!\in\!\mathbb{R}^{10\times 200}$).}
  We observe outliers such as $\approx-1.61$ and $\approx+2.56$ in anti-grokking.
\end{itemize}
These large coordinates coincide with pronounced increases of the
\texttt{WeightWatcher} trap statistic in the anti-grokking phase (cf.\ the trap
table in Sec.~\ref{app:modadd}, where mean traps rise from
$\approx 0$ in pre-/grokking to $\approx 4.13\pm1.13$ in anti-grokking).
Together with the sufficient condition in App.~\ref{app:shuffle_outliers}, these observations support the following chain:
\[
\text{Rank-1 perturbations} \;\Longrightarrow\;
\text{MP-outlying eigenvalues} \;\Longrightarrow\; \text{trap eigenpairs after shuffling}.
\]

\paragraph{From eigenvalues to vectors: what the largest tail vectors look like.}
Write the thin SVD $W=U\Sigma V^{\!\top}$, so the eigenpairs of $X(W)$ are
$\bigl(\sigma_k^{2}/N,\; v_k\bigr)$ with $v_k$ the $k$-th column of $V$.
Hence the \emph{tail vector} $v_1$ associated with
$\lambda_{\max}(W)=\sigma_1^2/N$ lies in the \emph{input space} of $W$.
This makes tail vectors directly interpretable:
\begin{itemize}
\item \textbf{Layer $W_1$.} $X(W_1)=(1/N)W_1^{\!\top}W_1$ acts on pixels.
  The top right singular vector $v^{(1)}\!\in\!\mathbb{R}^{784}$ is a pixel-space
  pattern. Empirically, $v^{(1)}$ evolves from
  (i) unstructured noise at pregrokking, to
  (ii) a smooth, global ring-like template at best-test (grokking), to
  (iii) \emph{digit-shaped} templates (looks like an 8) in anti-grokking phase.
  The last stage indicates a strong localized structure consistent with \emph{memorization}.
\item \textbf{Layer $W_2$.} $X(W_2)$ acts on the \emph{output of layer 1}.
  Its top right singular vector $v^{(2)}\!\in\!\mathbb{R}^{200}$ identifies the
  layer-1 neurons most involved in the dominant correlated mode.
  Back-projecting $v^{(2)}$ through $W_1$ to pixel space,
  $v^{(2)}_{\mathrm{pix}} := v^{(2)\!\top}\! W_1\!\in\!\mathbb{R}^{784}$,
  produces images that, in anti-grokking, again exhibit crisp digit strokes.
  The largest coefficients of $v^{(2)}$ pick out specific rows of $W_1$, and the
  corresponding receptive fields (rows of $W_1$ reshaped to $28\!\times\!28$)
  show clear digit templates in anti-grokking, whereas they are noise-like at
  pre-grok and remain largely uninterpretable at best-test.
\item \textbf{Layer $W_3$.} $X(W_3)$ acts on the \emph{output of layer 2}.
  Its top right singular vector $v^{(3)}\!\in\!\mathbb{R}^{200}$ highlights the
  most excited layer-2 units. Back-projecting successively through $W_2$ and $W_1$,
  \[
  v^{(3)}_{\mathrm{pix}}
  \;:=\; v^{(3)\!\top}\! W_2\, W_1 \;\in\; \mathbb{R}^{784},
  \]
  yields pixel-space patterns that sharpen into digit shapes in anti-grokking.
\end{itemize}

\paragraph{Mechanistic interpretation.}
In anti-grokking, you get non-random structural outliers; the ESD develops \emph{tail} spikes (large eigenvalues past this edge) , and the associated
eigenvectors localize onto coordinates resembling a sample or prototype being memorized from the train set. Entry-wise shuffling preserves the large rank-1 perturbations and thus the
outlying eigenvalues, converting these localized modes into \emph{trap} modes.
The empirical “rows-of-$W_1$’’ panels show this directly: noise-like at
pre-grokking phase, still diffuse at best-test, but in anti-grokking the top rows
look like \emph{clear digit templates}, evidencing \textbf{digit memorization}.

\paragraph{Practical procedure (used for the figures).}
For each checkpoint:
\begin{enumerate}\itemsep0.2em
\item Compute the SVD of each linear layer: $W_\ell=U_\ell\Sigma_\ell V_\ell^{\!\top}$.
  The \emph{tail vector} for layer $\ell$ is $v^{(\ell)}:=V_\ell[:,1]$.
\item For $W_1$ visualize $v^{(1)}$ in pixel space (reshape to $28\times28$).
\item For $W_2$ back-project $v^{(2)}$ to pixel space via
  $v^{(2)}_{\mathrm{pix}}=v^{(2)\!\top}W_1$ and display the top-$k$ contributing
  rows of $W_1$ chosen by the largest coordinates of $v^{(2)}$.
\item For $W_3$ back-project $v^{(3)}$ through $W_2$ and $W_1$ to obtain
  $v^{(3)}_{\mathrm{pix}}$, and again inspect the top contributors.
\item In parallel, compute $\lambda_{\max}(X(W_\ell))$ and
  $\lambda_{\max}(X(W_{\ell,\mathrm{rand}}))$ and compare to $\lambda_{+}$ to
  flag \emph{trap} eigenpairs; relate these to the \texttt{WeightWatcher} trap
  counts and to the HTSR $\alpha$ estimates.
\end{enumerate}

\FloatBarrier
\begin{figure*}[t]
  \centering
  \begin{subfigure}[b]{0.50\textwidth}
    \centering
    \includegraphics[width=\linewidth]{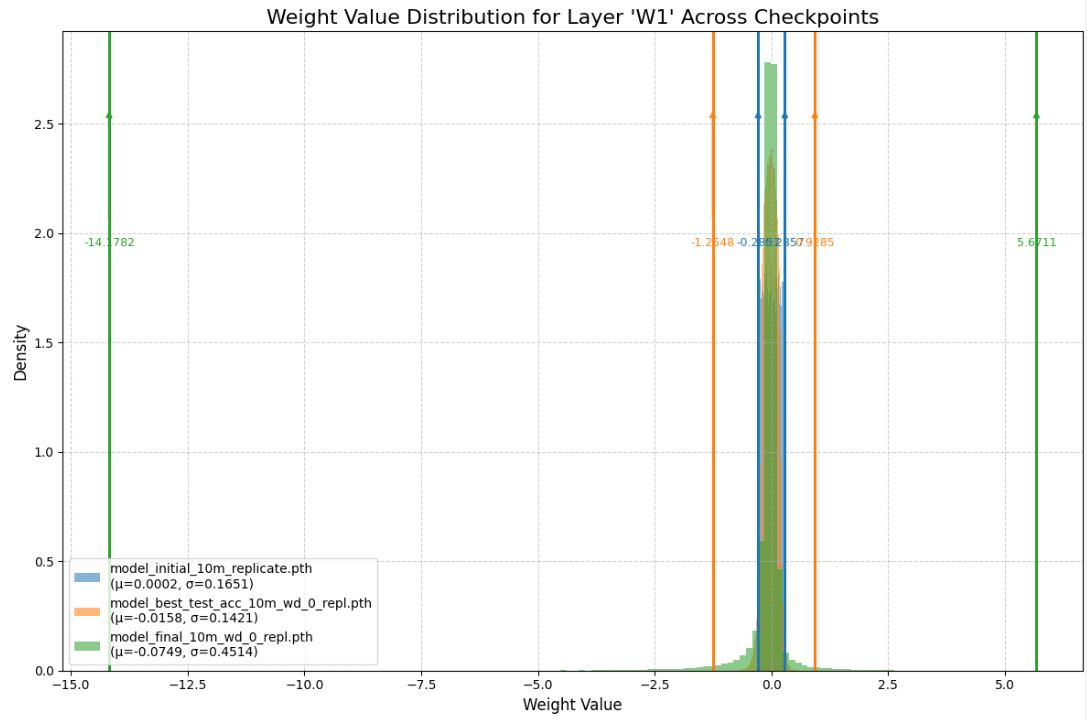}
    \caption{W1 weights across checkpoints}
    \label{fig:w1-hist}
  \end{subfigure}\hfill
  \begin{subfigure}[b]{0.50\textwidth}
    \centering
    \includegraphics[width=\linewidth]{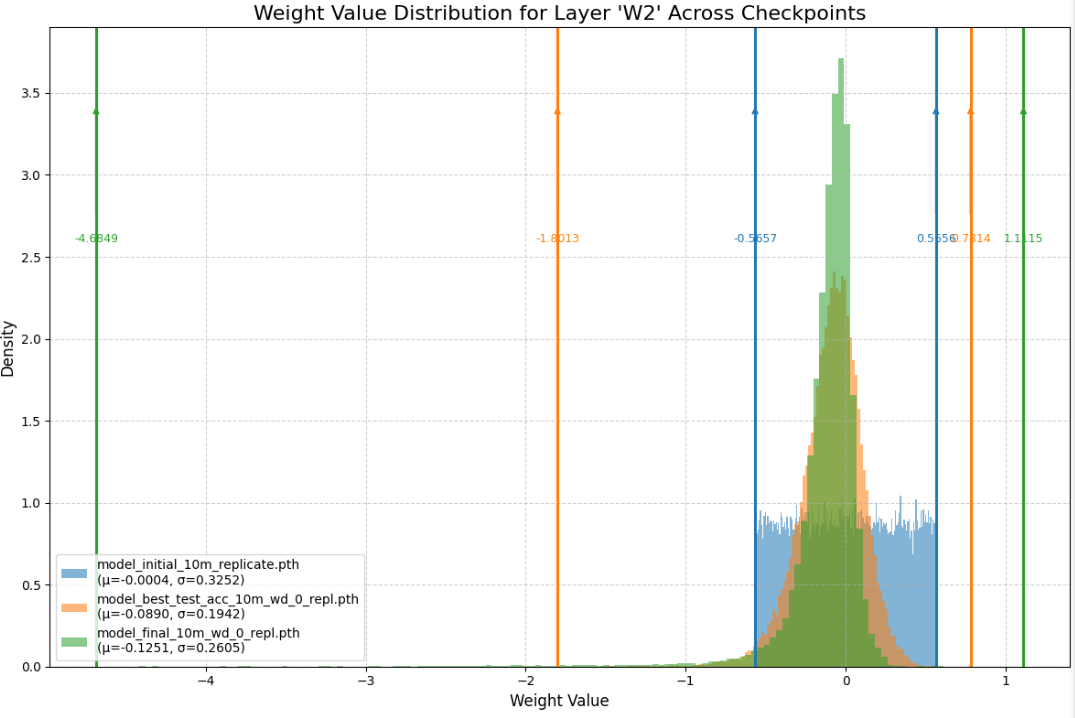}
    \caption{W2 weights across checkpoints}
    \label{fig:w2-hist}
  \end{subfigure}\hfill
  \begin{subfigure}[b]{0.50\textwidth}
    \centering
    \includegraphics[width=\linewidth]{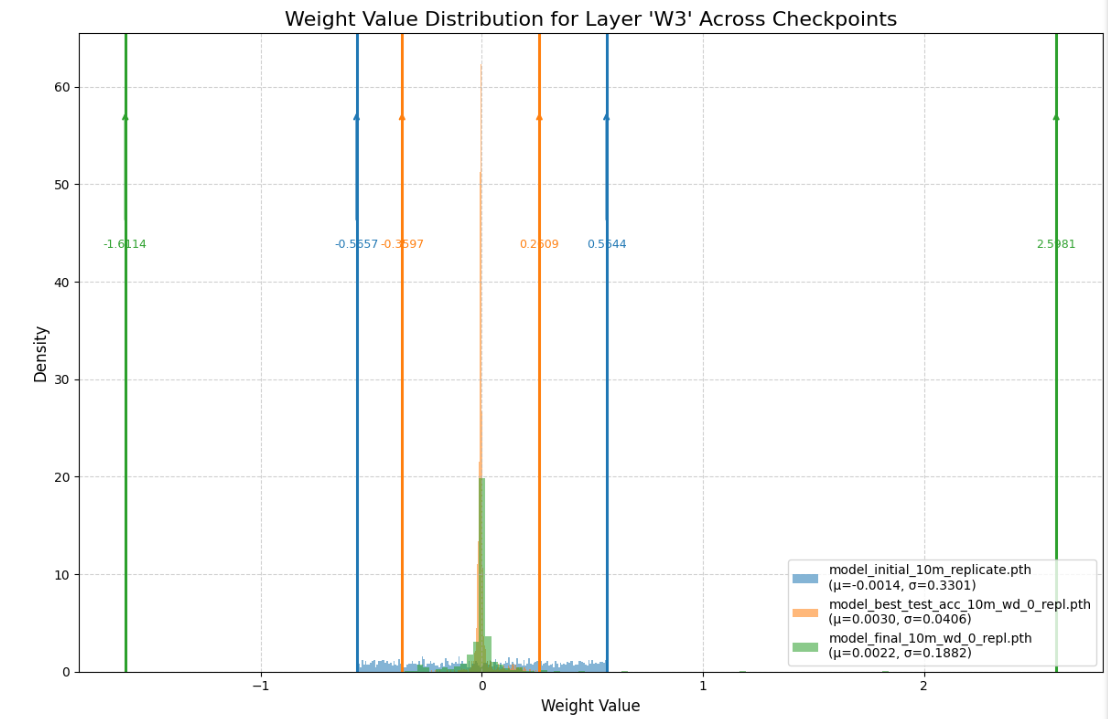}
    \caption{W3 weights across checkpoints}
    \label{fig:w3-hist}
  \end{subfigure}
  \caption{\textbf{Weight distributions with extreme elements.} Note in figures (a),(b) and (c) we see that there are structural outliers (rank-1 perturbations) causing the spikes in the randomized ESD (Correlation Traps)
 }
  \label{fig:w-hists}
\end{figure*}
\FloatBarrier

\begin{figure*}[t]
  \centering
  \begin{subfigure}[b]{1.00\textwidth}
    \centering
    \includegraphics[width=\linewidth]{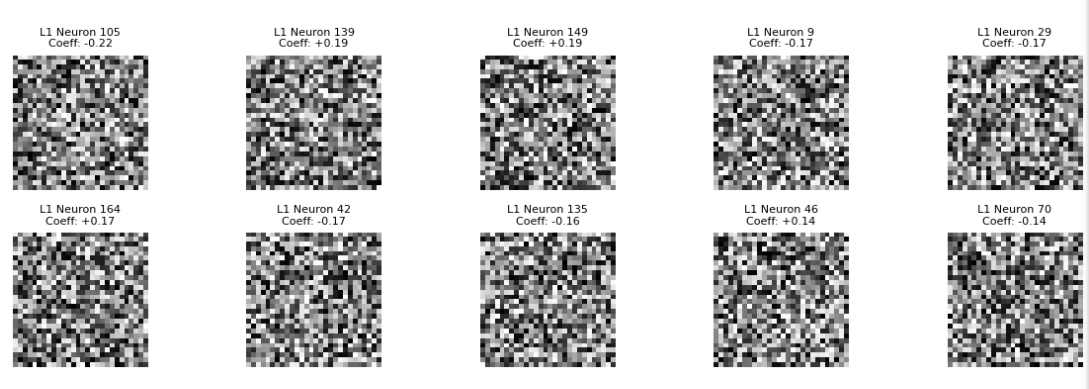}
    \caption{Top-10 rows of $W_1$ (pregrok)}
  \end{subfigure}\hfill
  \begin{subfigure}[b]{1.00\textwidth}
    \centering
    \includegraphics[width=\linewidth]{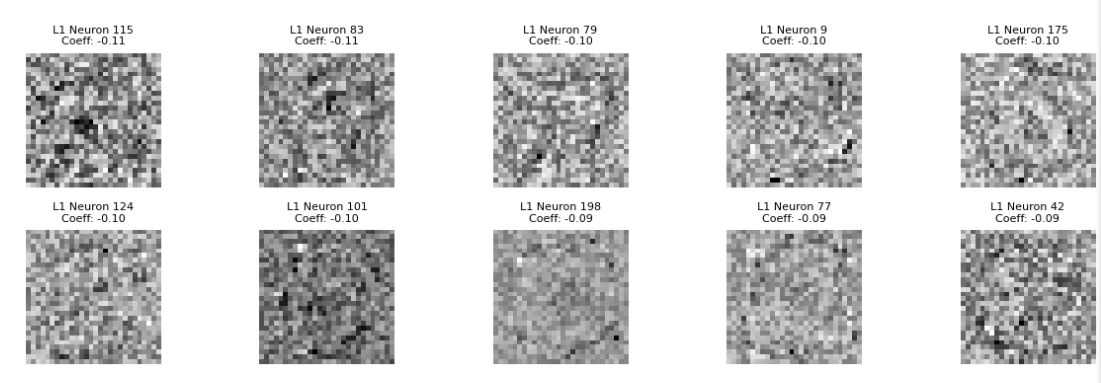}
    \caption{Top-10 rows of $W_1$ (grok)}
  \end{subfigure}\hfill
  \begin{subfigure}[b]{1.00\textwidth}
    \centering
    \includegraphics[width=\linewidth]{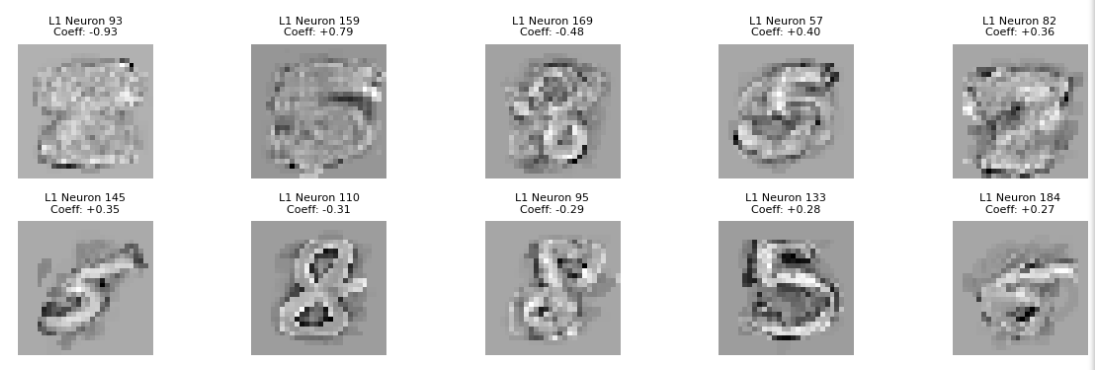}
    \caption{Top-10 rows of $W_1$ (anti-grok)}
  \end{subfigure}
  \caption{\textbf{Rows of $W_1$ (receptive fields) selected by tail vectors localization.}
  (a) shows noise phenotype at pregrok
, still diffuse at the grokking phase, but in anti-grokking the top rows
look like \emph{clear digit templates}, evidencing \textbf{digit memorization}.}
  \label{fig:l1-rfs}
\end{figure*}
\FloatBarrier

\section{Further Experiments: Modular Addition (predicting $x+y \bmod P$)}
\label{app:modadd}

We additionally evaluated the HTSR diagnostics on a modular-addition task
(predicting $x+y \bmod P$) with a small one-layer Transformer.
The accuracy curves again exhibit the three characteristic phases
(pre-grokking $\rightarrow$ grokking $\rightarrow$ anti-grokking), and both the
HTSR exponent $\alpha$ and \emph{Correlation Traps} (detected by
\texttt{WeightWatcher}) flag the late-stage generalization collapse.

\paragraph{Model architecture.}
\vspace{-0.5em}
\begin{table}[h!]
\centering
\small
\begin{tabular}{@{}ll@{}}
\toprule
\textbf{Hyper-parameter} & \textbf{Value} \\
\midrule
Layers & 1 \\
Model dimension $d_{\text{model}}$ & 128 \\
MLP hidden size $d_{\text{mlp}}$ & 512 \\
Heads $\times$ head dim & $4 \times 32$ \\
Context length $n_{\text{ctx}}$ & 3 \\
Activation & \textbf{ReLU} \\
LayerNorm & \emph{disabled} (\texttt{use\_ln=False}) \\
Vocabulary size & 114 (=$\,$\texttt{equals\_token}$\,+\,1$) \\
\bottomrule
\end{tabular}
\end{table}
\vspace{-1.0em}

As predicted by HTSR, the PL-exponent $\alpha$ concentrates near~2
around the peak test accuracy and then deviates from~2 thereafter. Consistent with
our main results, we also observe an increase in outlier eigenvalues escaping the
Marchenko–Pastur bulk in multiple layers during the anti-grokking phase.

\subsection{Phase and Layer-wise HTSR \texorpdfstring{$\alpha$}{alpha} for Modular Addition}
\label{app:phase_layer_tables_modadd}

\begin{table}[h!]
\centering
\caption{\textbf{Modular addition, phase summary.} Train and test accuracy (mean~$\pm$~std).}
\vspace{0.25em}
\label{tab:phase_summary_qn}
\begin{tabular}{@{}lcc@{}}
\toprule
\textbf{Phase} & \textbf{Train Acc.\ (mean$\pm$std)} & \textbf{Test Acc.\ (mean$\pm$std)} \\
\midrule
Pre-grok & $1.0\pm0.0$ & $0.40\pm0.28$ \\
Grok     & $1.0\pm0.0$ & $0.97\pm0.02$ \\
Anti-grok & $1.0\pm0.0$ & $0.68\pm0.11$ \\
\bottomrule
\end{tabular}
\end{table}

\begin{table}[h!]
\centering
\caption{\textbf{Modular addition, layer-wise HTSR $\alpha$ across phases.}
Means per layer; last row reports mean$\pm$std across listed layers.}
\vspace{0.25em}
\label{tab:layer_alpha_qn}
\begin{tabular}{@{}lccc@{}}
\toprule
\textbf{Layer} & \textbf{Pre-grok} & \textbf{Grok} & \textbf{Anti-grok} \\
\midrule
embed.embed             & 2.81 & 1.92 & 3.57 \\
blocks.0.attn.W\_Q      & 4.04 & 2.86 & 3.17 \\
blocks.0.attn.W\_K      & 3.99 & 2.00 & 3.20 \\
blocks.0.attn.W\_V      & 4.66 & 1.56 & 3.87 \\
blocks.0.attn.out\_proj & 3.85 & 1.57 & 4.04 \\
blocks.0.mlp.fc1        & 4.99 & 1.71 & 4.74 \\
blocks.0.mlp.fc2        & 5.22 & 2.17 & 5.03 \\
unembed.unembed         & 3.27 & 2.37 & 3.53 \\
\midrule
\textbf{Mean$\pm$Std}   & $\mathbf{4.10\pm0.83}$ & $\mathbf{2.02\pm0.44}$ & $\mathbf{3.89\pm0.68}$ \\
\bottomrule
\end{tabular}
\end{table}

\begin{table}[h!]
\centering
\caption{\textbf{Modular addition, correlation-trap counts.} Number of outlier eigenvalues (beyond the MP edge) detected by \texttt{WeightWatcher} in $X(W^{\mathrm{rand}})$ for each layer and phase.}
\vspace{0.25em}
\label{tab:layer_traps_qn}
\begin{tabular}{@{}lccc@{}}
\toprule
\textbf{Layer} & \textbf{Pre-grok traps} & \textbf{Grok traps} & \textbf{Anti-grok traps} \\
\midrule
embed.embed             & 0.00 & 0.00 & 7 \\
blocks.0.attn.W\_Q      & 0.00 & 0.00 & 3 \\
blocks.0.attn.W\_K      & 0.00 & 0.00 & 5 \\
blocks.0.attn.W\_V      & 0.00 & 0.00 & 3 \\
blocks.0.attn.out\_proj & 0.00 & 0.00 & 4 \\
blocks.0.mlp.fc1        & 0.00 & 0.00 & 5 \\
blocks.0.mlp.fc2        & 0.00 & 0.00 & 7 \\
unembed.unembed         & 0.00 & 0.00 & 5 \\
\midrule
\textbf{Mean$\pm$Std}   & $\mathbf{0.00\pm0.00}$ & $\mathbf{0.00\pm0.00}$ & $\mathbf{4.88\pm1.55}$ \\
\bottomrule
\end{tabular}
\end{table}

\paragraph{Interpretation.}
During grokking, most layers approach the \emph{ideal} regime $\alpha\!\approx\!2$,
indicating strong, well-structured correlations and high generalization.
In the anti-grokking phase, the average $\alpha$ drifts upward again while
\emph{selected layers} exhibit $\alpha<2$ and clear correlation-trap outliers
in their randomized spectra $X(W^{\mathrm{rand}})$ (beyond the MP edge), as
flagged automatically by \texttt{WeightWatcher}. This reproduces, on a distinct
algorithmic task, the same qualitative signature reported for MNIST:
$\alpha$ tracks all three phases, and the joint condition \emph{$\alpha\neq2$ + traps}
reliably marks late-stage generalization collapse.  \emph{Note that many of the layers in the anti-grokking phase are severely atypical, with multiple Correlation Traps, rank collapse, and even multi-model power-law signatures; for this reason, the anti-grokking $\alpha$ may be unreliable.}

% \end{appendices}

% Requires in preamble:
% \usepackage{graphicx}
% \usepackage{booktabs}
% \usepackage{amsmath,amssymb}
% \usepackage{array}

\section*{Appendix H \quad Modular Addition: Three-phase Empirical Analysis (PRE, GROK, ANTI)}

\subsection*{H.0 \quad Measurement protocol and computed quantities}

All computations in this appendix are performed on the above modular-addition task with modulus $p=113$.
Numeric tokens are $\{0,\dots,112\}$ and the equals token has index $113$.
The model is a single-layer decoder-only transformer with $D_{\text{model}}=128$, $D_{\text{MLP}}=512$, $N_{\text{heads}}=4$, $D_{\text{head}}=32$, context length $3$, ReLU activations, and no LayerNorm.
For each phase label $L\in\{\text{PRE},\text{GROK},\text{ANTI}\}$ the following objects are computed.

\paragraph{(1) Keys, query, Gram spectra, and preference scores.}
For each head $h\in\{0,1,2,3\}$:
\begin{align*}
K_0[x] &\in \mathbb{R}^{D_{\text{head}}} \quad \text{from the position-$0$ key on the length-3 input } [x,0,=],\\
K_1[y] &\in \mathbb{R}^{D_{\text{head}}} \quad \text{from the position-$1$ key on } [0,y,=],\\
q_{\mathrm{eq}} &\in \mathbb{R}^{D_{\text{head}}} \quad \text{from the equals-position query on } [0,0,=].
\end{align*}
Let $K^\circ[t]=K[t]-\frac{1}{p}\sum_{u=0}^{p-1}K[u]$ denote token-centering.
We then define the Correlation matrices
\[
G_0=\frac{1}{D_{\text{head}}}K_0^\circ\,(K_0^\circ)^\top,\qquad
G_1=\frac{1}{D_{\text{head}}}K_1^\circ\,(K_1^\circ)^\top,
\]
and report the largest eigenvalues $\lambda_{\max}(G_0)$ and $\lambda_{\max}(G_1)$.
Preference magnitudes use
\[
s_0[x]=\frac{\langle K_0[x],\,q_{\mathrm{eq}}\rangle}{\sqrt{D_{\text{head}}}},\qquad
s_1[y]=\frac{\langle K_1[y],\,q_{\mathrm{eq}}\rangle}{\sqrt{D_{\text{head}}}},
\]
with the top-5 token indices listed by $|s_0|$ and $|s_1|$.

\paragraph{(2) Token–DFT energy profiles.}
For any token-by-feature matrix $M\in\mathbb{R}^{p\times d}$ (here $M\in\{\texttt{Embed},\texttt{Unembed},K_0\text{(concat)},K_1\text{(concat)}\}$), let
\[
W_{f,t}=\frac{1}{\sqrt{p}}\,e^{-2\pi i f t/p}\quad(0\le f,t<p),\qquad
F=WM,\qquad \text{power}(f)=\frac{1}{d}\sum_{j=1}^{d}\bigl|F_{f,j}\bigr|^2.
\]
The normalized energy is $\mathsf{E}_M(f)=\text{power}(f)/\sum_{g}\text{power}(g)$.
We report the six largest frequency indices (``top idx''), their normalized values (``vals''), and the non-DC mass $\sum_{f=1}^{p-1}\mathsf{E}_M(f)$.

\paragraph{(3) Rule kernel over $\Delta=z-(x+y)$.}
From logits $L(x,y,z)$ (restricted to $z\in\{0,\dots,112\}$), define
\[
k(\Delta)\;=\;\frac{1}{p^2}\sum_{x,y=0}^{p-1} L\!\left(x,y,\,(x+y+\Delta)\bmod p\right)\;-\;\frac{1}{p}\sum_{\Delta'=0}^{p-1}\frac{1}{p^2}\sum_{x,y=0}^{p-1}L\!\left(x,y,\,(x+y+\Delta')\bmod p\right).
\]
We list the top-6 $\Delta$ by $k(\Delta)$.
We also compute $\widehat{k}=Wk$ and report the top-6 frequency indices by $|\widehat{k}(f)|^2$ and their powers.

\medskip

\subsection*{H.1 \quad Summary of accuracies and global set relations}

\begin{table}[h!]
\centering
\caption{Accuracies and global frequency-set relations.
``Kernel = Emb/Unemb freqs'' indicates whether the kernel’s top-6 DFT indices match those of \texttt{Embed}/\texttt{Unembed}.
``$K_0/K_1$ top-6 contains DC'' indicates whether index $0$ appears among the top-6 for $K_0$/$K_1$.
The set $S=\{49,64,97,16,6,107\}$ is used for intersection counts.}
\label{tab:acc-global}
\begin{tabular}{lcccc}
\toprule
Phase & Train Acc. & Test Acc. & Kernel = Emb/Unemb freqs & $K_0/K_1$ top\textendash6 contains DC (0) \\
\midrule
PRE  & $1.0$ & $0.13$ & Yes & Yes / Yes \\
GROK & $1.0$ & $1.0$                & Yes & Yes / Yes \\
ANTI & $1.0$ & $0.60$ & Yes & No  / No  \\
\bottomrule
\end{tabular}

\vspace{6pt}
\begin{tabular}{lcc}
\toprule
Phase & $|S\cap \text{top-6}(K_0)|$ & $|S\cap \text{top-6}(K_1)|$ \\
\midrule
PRE  & $2$ & $2$ \\
GROK & $5$ & $5$ \\
ANTI & $6$ & $6$ \\
\bottomrule
\end{tabular}
\end{table}

\subsection*{H.2 \quad Per-head keys, Gram largest eigenvalues, and preference indices}

\begin{table}[h!]
\centering
\caption{Per-head $\lambda_{\max}(G_0)$, $\lambda_{\max}(G_1)$, and top-5 token indices by $|\langle q_{\mathrm{eq}},K\rangle|$.}
\label{tab:keys-prefs}
\small
\begin{tabular}{llccp{0.36\textwidth}p{0.36\textwidth}}
\toprule
Phase & Head & $\lambda_{\max}(G_0)$ & $\lambda_{\max}(G_1)$ & Top-5 for $|{\langle q,K_0\rangle}|$ & Top-5 for $|{\langle q,K_1\rangle}|$ \\
\midrule
\multirow{4}{*}{PRE}
& 0 & 0.86 & 0.86 & [73, 55, 38, 34, 91] & [76, 30, 9, 7, 85] \\
& 1 & 1.07 & 1.07 & [63, 79, 1, 24, 40]  & [110, 32, 41, 11, 82] \\
& 2 & 1.61 & 1.61 & [76, 83, 84, 10, 73] & [83, 84, 10, 73, 51] \\
& 3 & 2.09 & 2.09 & [96, 18, 22, 110, 44] & [76, 41, 81, 40, 96] \\
\midrule
\multirow{4}{*}{GROK}
& 0 & 1.32 & 1.32 & [19, 40, 5, 111, 26]  & [19, 40, 5, 111, 26] \\
& 1 & 2.05 & 2.05 & [78, 101, 108, 2, 85] & [78, 101, 108, 2, 85] \\
& 2 & 1.56 & 1.56 & [72, 65, 86, 79, 102] & [72, 65, 86, 79, 102] \\
& 3 & 2.19 & 2.19 & [73, 66, 50, 80, 43]  & [73, 66, 50, 80, 43] \\
\midrule
\multirow{4}{*}{ANTI}
& 0 & 123.85 & 123.85 & [75, 98, 29, 45, 31] & [75, 98, 29, 45, 31] \\
& 1 & 98.89  & 98.89  & [71, 3, 63, 70, 94]  & [71, 3, 63, 70, 94] \\
& 2 & 123.01 & 123.01 & [72, 5, 56, 93, 6]   & [72, 5, 56, 93, 6] \\
& 3 & 94.14  & 94.14  & [4, 80, 58, 11, 103] & [4, 80, 11, 58, 103] \\
\bottomrule
\end{tabular}
\end{table}

\subsection*{H.3 \quad Token–DFT energy (dominant indices and non-DC mass)}

\begin{table}[h!]
\centering
\caption{Top-6 DFT and normalized energies for the token-indexed matrices, with non-DC mass. Indices are in $\{0,\dots,112\}$.}
\label{tab:fourier-energy}
\small
\begin{tabular}{llp{0.34\textwidth}p{0.34\textwidth}c}
\toprule
Phase & Module & Top idx (in reported order) & Values for top idx (same order) & non-DC mass \\
\midrule
\multirow{4}{*}{PRE}
& Embed   & [49, 64, 97, 16, 6, 107] & [0.0667, 0.0667, 0.0509, 0.0509, 0.0200, 0.0200] & 1.000 \\
& Unembed & [49, 64, 97, 16, 6, 107] & [0.0714, 0.0714, 0.0484, 0.0484, 0.0222, 0.0222] & 1.000 \\
& $K_0$   & [0, 104, 9, 64, 49, 20]  & [0.6845, 0.0180, 0.0180, 0.0130, 0.0130, 0.0080] & 0.315 \\
& $K_1$   & [0, 104, 9, 64, 49, 20]  & [0.7011, 0.0170, 0.0170, 0.0123, 0.0123, 0.0076] & 0.299 \\
\midrule
\multirow{4}{*}{GROK}
& Embed   & [49, 64, 97, 16, 6, 107] & [0.2181, 0.2181, 0.1270, 0.1270, 0.1015, 0.1015] & 1.000 \\
& Unembed & [49, 64, 97, 16, 6, 107] & [0.2144, 0.2144, 0.1438, 0.1438, 0.1405, 0.1405] & 1.000 \\
& $K_0$   & [49, 64, 0, 97, 16, 6]   & [0.3660, 0.3660, 0.1076, 0.0757, 0.0757, 0.0039] & 0.892 \\
& $K_1$   & [49, 64, 0, 97, 16, 6]   & [0.3658, 0.3658, 0.1081, 0.0757, 0.0757, 0.0039] & 0.892 \\
\midrule
\multirow{4}{*}{ANTI}
& Embed   & [49, 64, 97, 16, 6, 107] & [0.0210, 0.0210, 0.0206, 0.0206, 0.0199, 0.0199] & 0.992 \\
& Unembed & [49, 64, 97, 16, 6, 107] & [0.0264, 0.0264, 0.0218, 0.0218, 0.0201, 0.0201] & 0.992 \\
& $K_0$   & [16, 97, 107, 6, 49, 64] & [0.0260, 0.0260, 0.0230, 0.0230, 0.0224, 0.0224] & 0.985 \\
& $K_1$   & [16, 97, 107, 6, 49, 64] & [0.0260, 0.0260, 0.0230, 0.0230, 0.0224, 0.0224] & 0.985 \\
\bottomrule
\end{tabular}
\end{table}

\subsection*{H.4 \quad Rule kernel $k(\Delta)$ and its discrete Fourier spectrum}

\begin{table}[h!]
\centering
\caption{Top-6 $\Delta$ for $k(\Delta)$ and the top-6 DFT frequencies of $k$ with powers $|\widehat{k}(f)|^2$.}
\label{tab:kernel}
\small
\begin{tabular}{l p{0.34\textwidth} p{0.34\textwidth}}
\toprule
Phase & Top-6 $\Delta$ (in reported order) & Top-6 DFT freqs (idx) with powers (same order) \\
\midrule
PRE  & [0, 7, 106, 14, 99, 21] & idx [49, 64, 97, 16, 6, 107];\; pow [88{,}713.516, 88{,}713.516, 8{,}677.078, 8{,}677.078, 61.569, 61.569] \\
GROK & [0, 37, 76, 92, 21, 99] & idx [49, 64, 97, 16, 6, 107];\; pow [208{,}933.61, 208{,}933.61, 10{,}937.356, 10{,}937.356, 4{,}895.602, 4{,}895.602] \\
ANTI & [0, 21, 92, 76, 37, 99] & idx [49, 64, 97, 16, 6, 107];\; pow [9{,}955{,}823.0, 9{,}955{,}823.0, 1{,}597{,}666.1, 1{,}597{,}666.1, 1{,}207{,}065.4, 1{,}207{,}065.4] \\
\bottomrule
\end{tabular}

\vspace{6pt}
\noindent\textit{Note on index pairs.} With $p=113$, the frequency indices occur in complementary pairs $(f,\,113-f)$; the sets $\{49,64\}$, $\{97,16\}$, $\{6,107\}$ satisfy $64=113-49$, $97=113-16$, and $107=113-6$.
\end{table}

\medskip
\noindent\textbf{Cross-phase observations (data-derived).}
Across all phases, the dominant frequency indices for \texttt{Embed}, \texttt{Unembed}, and the kernel coincide with $S=\{49,64,97,16,6,107\}$.
For $K_0$ and $K_1$, the top-6 intersections with $S$ are $(2,2)$ in PRE, $(5,5)$ in GROK, and $(6,6)$ in ANTI.
The top-5 preference token lists for $|{\langle q_{\mathrm{eq}},K_0\rangle}|$ and $|{\langle q_{\mathrm{eq}},K_1\rangle}|$ are identical for every head in GROK; they differ on all heads in PRE; in ANTI they are identical on heads 0--2 and contain the same five tokens in a different order on head 3.

\subsection*{H.6 \quad Embedding/Unembedding Fourier profiles (overlay) and comparison across phases}

\begin{figure}[h!]
  \centering
  % Replace filenames below with your actual paths
  \includegraphics[width=.88\linewidth]{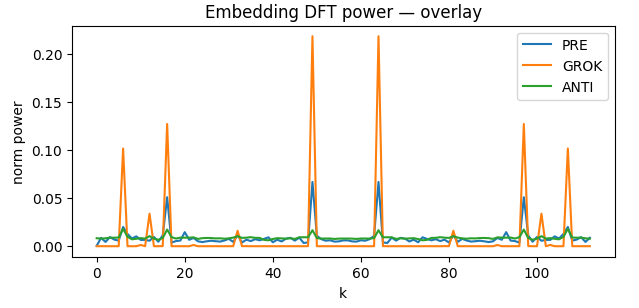}
  \caption{\textbf{Embedding DFT power (overlay).} Dominant nonzero frequencies are the complementary pairs
  $\{49,64\}$, $\{97,16\}$, $\{6,107\}$. The GROK trace shows sharp peaks; PRE shows smaller peaks;
  ANTI appears comparatively broadband with shallow peaks at the same indices.}
  \label{fig:embed-dft-overlay}
\end{figure}

\begin{figure}[h!]
  \centering
  % Replace filenames below with your actual paths
  \includegraphics[width=.88\linewidth]{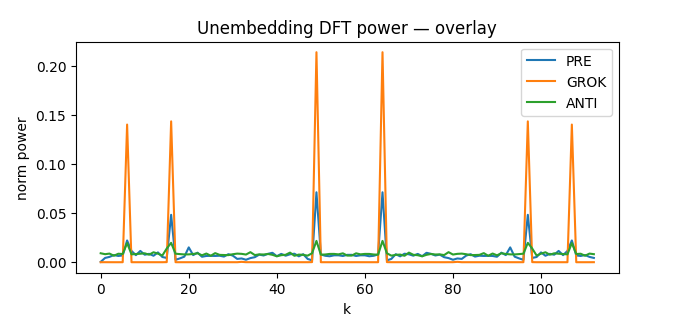}
  \caption{\textbf{Unembedding DFT power (overlay).} Same six nonzero frequencies as in the embedding.
  GROK is sharply peaked, PRE is weaker, and ANTI is comparatively broadband.}
  \label{fig:unembed-dft-overlay}
\end{figure}

Figures~\ref{fig:embed-dft-overlay}--\ref{fig:unembed-dft-overlay} plot the normalized token–DFT energy
$\mathsf{E}_M(f)$ for the \texttt{Embed} and \texttt{Unembed} matrices for the three phases.
The horizontal axis is the frequency index $k\in\{0,\dots,112\}$; the vertical axis is the normalized power
$\mathsf{E}_M(k)$ as defined in Section~H.0(2).

\vspace{4pt}
\noindent\textbf{Quantitative check (from Tables~\ref{tab:fourier-energy} and \ref{tab:kernel}).}
All three phases share the same six non-DC indices for \texttt{Embed} and \texttt{Unembed}: $[49,64,97,16,6,107]$.
Peak magnitudes at these indices differ:
\emph{Embed} peaks are $\approx 0.218$ (GROK), $\approx 0.067$ (PRE), $\approx 0.021$ (ANTI);
\emph{Unembed} peaks are $\approx 0.214$ (GROK), $\approx 0.071$ (PRE), $\approx 0.026$ (ANTI).
Hence GROK is highly concentrated on the six-rule frequency set, PRE shows the same set with smaller peaks, and ANTI exhibits a broadband profile (same indices but low peak-to-background ratios).

\vspace{4pt}
\noindent\textbf{Relation to key banks.}
For ANTI, $K_0$ and $K_1$ are not broadband: their top-6 frequency indices are exactly the six non-DC rule indices with non-DC mass $0.985$ (Table~\ref{tab:fourier-energy}).
Thus ANTI combines keys tightly supported on the rule frequencies with embeddings/unembedding that are comparatively broadband.
This mismatch is visible in Figures~\ref{fig:embed-dft-overlay}--\ref{fig:unembed-dft-overlay}.

\paragraph*{Interpretation}
\begin{itemize}
  \item From the per-head preference lists and Gram spectra (Table~9), \textbf{PRE} lacks the position-symmetry expected of modular addition; \textbf{GROK} exhibits exact symmetry; \textbf{ANTI} exhibits the same symmetry with much larger key-bank variance.
  \item Therefore, \textbf{ANTI}'s failure cannot be attributed to attention key--query architecture; it must arise elsewhere as we show below.
  
  \item \textbf{Embedding/Unembedding Fourier profiles (Tables~\ref{tab:fourier-energy} and \ref{tab:kernel}, Figs.~\ref{fig:embed-dft-overlay}--\ref{fig:unembed-dft-overlay}).}
  All phases share the same six non-DC indices \(\{49,64,97,16,6,107\}\) for \texttt{Embed} and \texttt{Unembed}, but the peak magnitudes differ sharply:
  \emph{Embed} peaks are approximately: \\  
\hspace*{2em}\textbf{GROK}: \([0.2181,\,0.2181,\,0.1270,\,0.1270,\,0.1015,\,0.1015]\)\\
\hspace*{2em}\textbf{PRE}:  \([0.0667,\,0.0667,\,0.0509,\,0.0509,\,0.0200,\,0.0200]\)\\
\hspace*{2em}\textbf{ANTI}: \([0.0210,\,0.0210,\,0.0206,\,0.0206,\,0.0199,\,0.0199]\)  

\emph{Unembed} peaks show the same ordering: \\  
\hspace*{2em}\textbf{GROK}: \([0.2144,\,0.2144,\,0.1438,\,0.1438,\,0.1405,\,0.1405]\)\\
\hspace*{2em}\textbf{PRE}:  \([0.0714,\,0.0714,\,0.0484,\,0.0484,\,0.0222,\,0.0222]\)\\
\hspace*{2em}\textbf{ANTI}: \([0.0264,\,0.0264,\,0.0218,\,0.0218,\,0.0201,\,0.0201]\).

Thus, while the six frequencies are present in all phases, \textbf{ANTI} exhibits a comparatively \emph{broadband} \texttt{Embed}/\texttt{Unembed} profile (low peak-to-background), in contrast to the sharp, high-amplitude lines in \textbf{GROK}.
 
\item \textbf{Kernel vs.\ embedding contrast.}
The rule kernel \(k(\Delta)\) has the same six-frequency support in all phases, and its top-line powers are largest in \textbf{ANTI}: \\  
\hspace*{2em}\textbf{ANTI}: \([9955823.0,\,9955823.0,\,1597666.1,\,1597666.1,\,1207065.4,\,1207065.4]\)\\
\hspace*{2em}\textbf{GROK}: \([208933.61,\,208933.61,\,10937.356,\,10937.356,\,4895.602,\,4895.602]\)\\
\hspace*{2em}\textbf{PRE}:  \([88713.516,\,88713.516,\,8677.078,\,8677.078,\,61.569,\,61.569]\)  

Hence, even though the rule kernel (and the keys) are strongly aligned to the six-rule frequencies in \textbf{ANTI}, the \texttt{Embed}/\texttt{Unembed} spectra \emph{lose the sharp Fourier spikes} and appear broadband.

\end{itemize}

\FloatBarrier
\begin{figure*}[t]
  \centering

  % -------- row 1: PRE-GROK --------
  \begin{subfigure}[t]{0.48\textwidth}
    \centering
    \includegraphics[width=\linewidth]{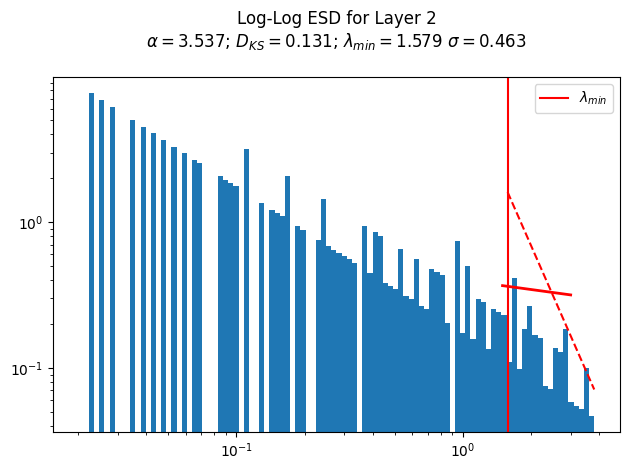}
    \caption{Pre-grok: embed ESD vs.\ randomized}
    \label{fig:esd-pre}
  \end{subfigure}\hfill
  \begin{subfigure}[t]{0.48\textwidth}
    \centering
    % rename if your filename differs
    \includegraphics[width=\linewidth]{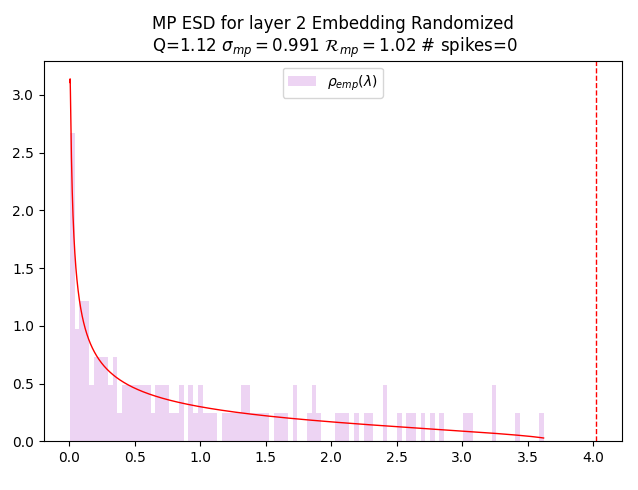}
    \caption{Pre-grok: MP fit \& bulk edge}
    \label{fig:esd-mp-pre}
  \end{subfigure}

  \vspace{0.8em} % space between rows

  % -------- row 2: GROK --------
  \begin{subfigure}[t]{0.48\textwidth}
    \centering
    \includegraphics[width=\linewidth]{weightwatcher_spike_embed_grok_2.png}
    \caption{Grok: embed ESD vs.\ randomized}
    \label{fig:esd-grok}
  \end{subfigure}\hfill
  \begin{subfigure}[t]{0.48\textwidth}
    \centering
    % rename if your filename differs
    \includegraphics[width=\linewidth]{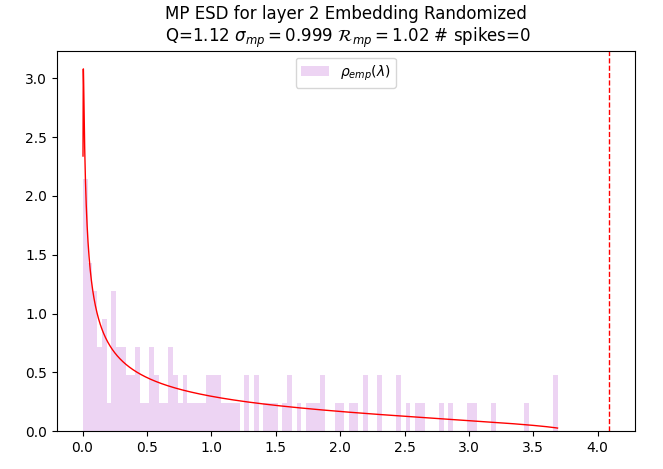}
    \caption{Grok: MP fit \& bulk edge}
    \label{fig:esd-mp-grok}
  \end{subfigure}

  \vspace{0.8em} % space between rows

  % -------- row 3: ANTI-GROK --------
  \begin{subfigure}[t]{0.48\textwidth}
    \centering
    \includegraphics[width=\linewidth]{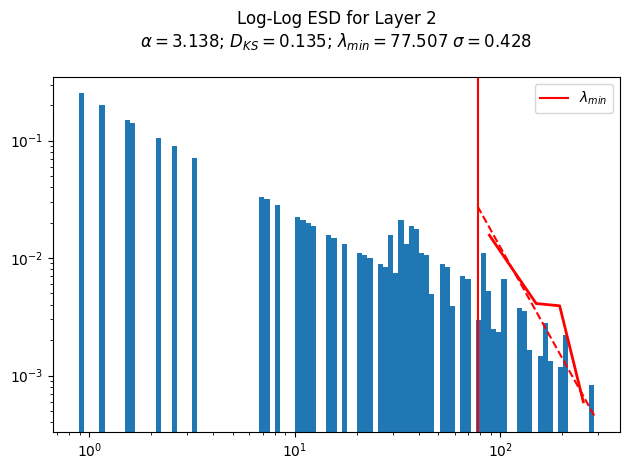}
    \caption{Anti-grok: embed ESD vs.\ randomized}
    \label{fig:esd-anti}
  \end{subfigure}\hfill
  \begin{subfigure}[t]{0.48\textwidth}
    \centering
    \includegraphics[width=\linewidth]{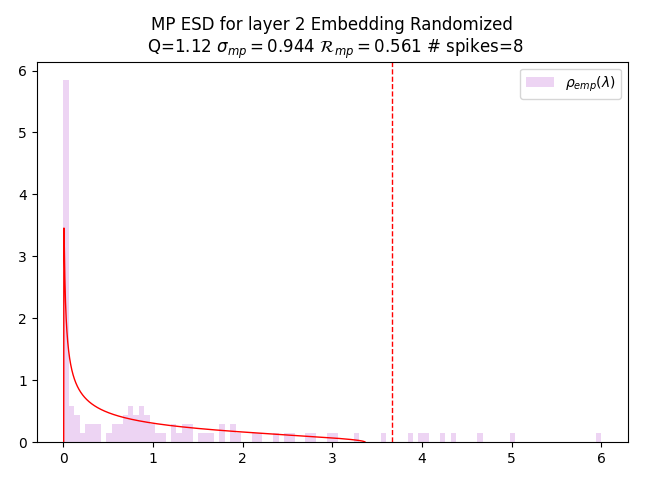}
    \caption{Anti-grok: MP fit \& bulk edge}
    \label{fig:esd-mp-anti}
  \end{subfigure}

  \caption{\textbf{Embedding-layer ESD across phases (3$\times$2 grid).}
  For each phase (rows), the left panel shows the empirical spectral density (ESD) for $\mathbf{W}_{\text{embed}}$ (blue) on a log-log scale, along with the best PL fit of the tail (red), and the associated $\alpha$ and other metrics for the fit, The right panel shows the Marchenko–Pastur (MP) fit of the ESD of the randomized matrix $\mathbf{W}_{\text{embed}}^{\text{rand}}$ and the far edge of the Tracy-Widom (TW) fluctuations (dotted vertical line). For all the Figures on the left, (a), (c), and (e), show distinct power law (PL) behavior across the entire ESD, not just the tail, which has its own, separate PL fit. This is highly unusual, and indicates strong overfitting even in the presence of correlated learning.  In the grokking phase, Figure (c) shows a very good PL fit of the tail, with $\alpha \approx 2$, and a much smaller $D_{KS}$ value than in the other 2 phases.  Coming to the right panel, for both the pre-grokking and grokking phases,
  in Figures (b) and (d),  the MP theory (red line) is a very good fit to the randomized ESDs (pink). In contrast, in Figure (f), in the anti-grokking phase , the randomized ESD (pink) is extremely atypical.  The MP fit  of the bulk region is not very good, there are several Correlation Traps (are eigenvalues extending far beyond the randomized MP bulk edge ($\lambda_{trap}\gg\lambda^{+}_{rand})$), and there is also significant rank collapse (larger density at zero than the MP theory predicts).  Here, anti-grokking appears to correspond to catastrophic forgetting.}
  \label{fig:modadd-embed-esd-panel}
\end{figure*}
\FloatBarrier

\section{Perturbation studies and our metrics}
\label{app:l2-perturbation}

We try modulating the rescaling alpha to 4 
we find in particular, even when the parameters are globally rescaled (a simple perturbation inspired by Omnigrok-style analyses), the Weightwatcher layer quality metrics track the grokking phases.

\paragraph{Protocol and phase definitions.}
The phase boundaries change as follows :
\emph{pre-grok} ($10^2$--$5\times10^4$ steps), \emph{grok} ($5\times10^4$--$5\times10^5$ steps), and \emph{anti-grok} ($>5\times10^5$ steps).
For each phase, we evaluate training/test accuracy and the global parameter $\ell_2$ norm.
We additionally report layer-wise heavy-tail exponents (WeightWatcher $\alpha$) and a layer-wise correlation-trap statistic.
All values are reported as mean $\pm$ standard deviation over the samples used within each phase.

\begin{table}[H]
  \centering
  \caption{\textbf{Accuracy by phase.} Mean $\pm$ std training and test accuracy within each phase.}
  \label{tab:app-acc-by-phase}
  \begin{tabular}{lcc}
    \toprule
    Phase & Train Accuracy (mean$\pm$std) & Test Accuracy (mean$\pm$std) \\
    \midrule
    Pre-grok  & $0.7298 \pm 0.0408$ & $0.4926 \pm 0.0187$ \\
    Grok      & $1.0000 \pm 0.0000$ & $0.8879 \pm 0.0051$ \\
    Anti-grok & $1.0000 \pm 0.0000$ & $0.6184 \pm 0.1086$ \\
    \bottomrule
  \end{tabular}
\end{table}

\begin{table}[H]
  \centering
  \caption{\textbf{Global $\ell_2$ norm by phase.} Mean $\pm$ std of the parameter $\ell_2$ norm within each phase.}
  \label{tab:app-l2-by-phase}
  \begin{tabular}{lc}
    \toprule
    Phase & WeightNorm (mean$\pm$std) \\
    \midrule
    Pre-grok  & $17.52 \pm 0.00$ \\
    Grok      & $16.80 \pm 0.22$ \\
    Anti-grok & $27.64 \pm 8.39$ \\
    \bottomrule
  \end{tabular}
\end{table}

\begin{table}[H]
  \centering
  \caption{\textbf{Layer-wise $\alpha$ across phases.}
  WeightWatcher heavy-tail exponent estimates ($\alpha$) summarized within each phase.
  Phase windows are: pre-grok ($10^2$--$5\times10^4$), grok ($5\times10^4$--$5\times10^5$), anti-grok ($>5\times10^5$).}
  \label{tab:app-alpha-by-phase}
  \begin{tabular}{lccc}
    \toprule
    Layer & Pre-grok mean & Grok mean & Anti-grok mean \\
    \midrule
    Layer 1 & $8.88$ & $2.59$ & $1.23$ \\
    Layer 2 & $3.84$ & $2.77$ & $1.49$ \\
    \midrule
    Mean$\pm$Std & $\mathbf{6.36 \pm 2.52}$ & $\mathbf{2.68 \pm 0.09}$ & $\mathbf{1.36 \pm 0.13}$ \\
    \bottomrule
  \end{tabular}
\end{table}

\begin{table}[H]
  \centering
  \caption{\textbf{Correlation-trap analysis (layer-wise).}
  A layer-wise correlation-trap statistic summarized by phase (mean $\pm$ std over samples in each phase).}
  \label{tab:app-corr-traps}
  \begin{tabular}{lccc}
    \toprule
    Layer & Pre-grok & Grok & Anti-grok \\
    \midrule
    1 & $0.00$ & $0.00$ & $7.00$ \\
    3 & $0.00$ & $0.00$ & $1.17$ \\
    5 & $0.00$ & $0.00$ & $5.50$ \\
    \midrule
    Mean$\pm$Std & $\mathbf{0.00 \pm 0.00}$ & $\mathbf{0.00 \pm 0.00}$ & $\mathbf{4.56 \pm 3.03}$ \\
    \bottomrule
  \end{tabular}
\end{table}

% \paragraph{Why $\ell_2$ can fail as a predictor (geometric intuition).}
% A helpful way to view this is geometric: the set of parameter vectors with a fixed $\ell_2$ norm lies on a (very high-dimensional) hypersphere.
% Generalization is not constant on this hypersphere, clearly many points with the same $\ell_2$ norm can correspond to qualitatively different functions, ranging from chance-level test accuracy to strong generalization.
% Consequently, even if a perturbation rescales parameters (moving radially between hyperspheres), $\ell_2$-based thresholds alone need not isolate the subset of solutions that generalize well.

\end{appendices}

\end{document}